\documentclass[letterpaper, 10 pt, conference]{ieeeconf}

\newif\ifreview
\reviewfalse

\IEEEoverridecommandlockouts

\usepackage[resetlabels]{multibib}
\usepackage{subfiles}
\usepackage{csquotes}
\usepackage{gensymb}
\usepackage{pifont}
\usepackage{textcomp}
\usepackage[table]{xcolor}
\usepackage{todonotes}
\usepackage{blindtext}
\usepackage{soul}
\usepackage{graphicx}
\usepackage{mathtools}
\usepackage{multirow}
\usepackage{flushend}
\usepackage{booktabs}
\usepackage{array}
\usepackage{tabularx}

\usepackage{amsmath}
\usepackage{amssymb}
\usepackage{amsfonts}
\usepackage{bm}
\usepackage{siunitx}
\usepackage{cancel}
\let\labelindent\relax
\usepackage{enumitem}
\usepackage{microtype}
\usepackage{xcolor}
\usepackage[breaklinks,colorlinks]{hyperref}
\usepackage{caption}
\usepackage{cite}
\usepackage[export]{adjustbox}
\usepackage[hang,flushmargin]{footmisc}
\usepackage{threeparttable}
\usepackage{cuted}
\newcites{S}{References}

\usepackage[capitalize]{cleveref}
\crefname{section}{Sec.}{Secs.}
\Crefname{section}{Section}{Sections}
\Crefname{table}{Table}{Tables}
\crefname{table}{Tab.}{Tabs.}

\definecolor{mygreen}{HTML}{00A64F}
\definecolor{myred}{HTML}{ED1B23}

\renewcommand{\eqref}[1]{Eq.~(\ref{#1})}
\newcommand{\figref}[1]{Fig.~\ref{#1}}

\newcolumntype{P}[1]{>{\centering\arraybackslash}p{#1}}
\newcommand{\net}{ForecastOcc}
\newcommand{\adloss}{Future State Alignment}
\newcommand{\short}{FSA}
\usepackage{booktabs}

\title{\LARGE \bf
ForecastOcc: Vision-based Semantic Occupancy Forecasting
}

\ifreview
  \author{Anonymous Author(s)%
}
\else
  \author{Riya Mohan$^{*}$, Juana Valeria Hurtado$^{*}$, Rohit Mohan$^{*}$, Abhinav Valada%
\thanks{$^*$Equal contribution. All authors are with the Department of Computer Science, University of Freiburg, Germany.}%
    \thanks{This work was funded by the German Research Foundation (DFG) Emmy Noether Program grant number 468878300. Rohit Mohan was supported by the Bosch Research collaboration on AI-driven automated driving.}%
}
\fi

\begin{document}

\maketitle
\thispagestyle{empty}
\pagestyle{empty}

\begin{abstract}
Autonomous driving requires forecasting both geometry and semantics over time to effectively reason about future environment states. Existing vision-based occupancy forecasting methods focus on motion-related categories such as static and dynamic objects, while semantic information remains largely absent. Recent semantic occupancy forecasting approaches address this gap but rely on past occupancy predictions obtained from separate networks. This makes current methods sensitive to error accumulation and prevents learning spatio-temporal features directly from images. In this work, we present \net{}, the first framework for vision-based semantic occupancy forecasting that jointly predicts future occupancy states and semantic categories. Our framework yields semantic occupancy forecasts for multiple horizons directly from past camera images, without relying on externally estimated maps. We evaluate \net{} in two complementary settings: multi-view forecasting on the Occ3D-nuScenes dataset and monocular forecasting on SemanticKITTI, where we establish the first benchmark for this task. We introduce the first baselines by adapting two 2D forecasting modules within our framework. Importantly, we propose a novel architecture that incorporates a temporal cross-attention forecasting module, a 2D-to-3D view transformer, a 3D encoder for occupancy prediction, and a semantic occupancy head for voxel-level forecasts across multiple horizons. Extensive experiments on both datasets show that \net{} consistently outperforms baselines, yielding semantically rich, future-aware predictions that capture scene dynamics and semantics critical for autonomous driving.  We make the code publicly available at \url{https://forecastocc.cs.uni-freiburg.de}
\end{abstract}

\section{Introduction}

Understanding the temporal evolution of 3D scenes is essential for reliable decision-making in autonomous driving. This requires forecasting not only the geometry of the environment but also the semantic categories of objects in the scene and how they change over time~\cite{sekkat2024amodalsynthdrive}. Recent advances in vision-based occupancy estimation have shifted the focus from predefined object categories to predicting spatial occupancy states~\cite{ma2024cotr,zhang2023occformer}. Building on this, vision-based occupancy forecasting has emerged as the task of predicting how these states evolve in the future directly from images. Such methods replace semantic categories with a coarse static–dynamic split. While useful for planning, this overlooks the semantic detail required to reason about scene elements and their interactions~\cite{mohan2024panoptic, mohan2025open}.

\begin{figure}
    \centering
    \includegraphics[width=0.9\linewidth]{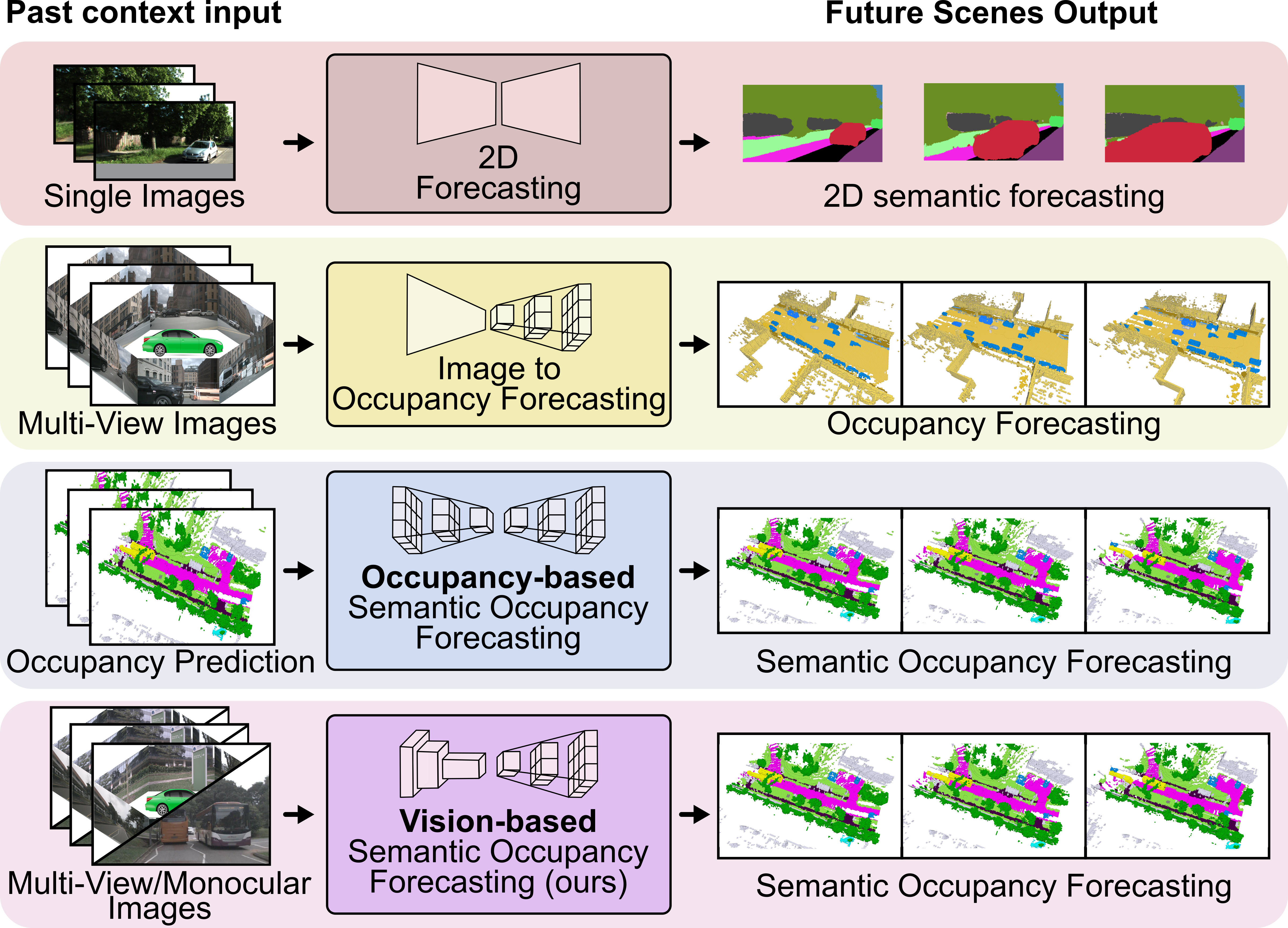}
    \caption{Different scene forecasting frameworks. Existing approaches include 2D semantic forecasting from single images which is limited to image-plane predictions. Image-based occupancy forecasting predicts future occupancy but lacks semantic understanding. Occupancy-based semantic forecasting requires occupancy predictions as input. In contrast, our vision-based semantic occupancy forecasting directly leverages multi-view images to jointly forecast both occupancy and semantics, enabling semantically rich 3D future scene understanding.}
    \label{fig:cover}
\end{figure}

In this direction, recent approaches have introduced semantic categories into occupancy forecasting~\cite{gu2024dome, liao20252, zheng2024occworld}. Instead of relying on motion-based categories, these methods forecast the future occupancy of surrounding agents and map elements such as drivable areas and vehicles, jointly with their semantic categories. However, current models for semantic occupancy forecasting depend on past 3D occupancy predictions as inputs. To adapt them to a vision-based setting, an additional network must be used to estimate current and past semantic occupancy maps. This introduces two main drawbacks. First, dependence on an external occupancy estimation model increases computational cost and risks error accumulation across stages. Second, separating semantic occupancy estimation from forecasting prevents joint learning of spatio-temporal features directly from raw camera inputs. As a result, these methods struggle to capture the full scene dynamics and semantic structure required for robust forecasting, as illustrated in Fig.~\ref{fig:cover}.

In this work, we introduce \net{}, a novel framework for vision-based semantic occupancy forecasting. Unlike prior approaches, \net{} takes camera images directly as input and jointly predicts future semantic occupancy states and categories without relying on externally estimated occupancy maps. \net{} delivers semantically rich, future-aware predictions that jointly capture the spatial, temporal, and semantic structure of future scenes.
We evaluate our approach in two complementary settings. First, we consider multi-view forecasting on the Occ3D-nuScenes dataset, where synchronized multi-view images provide broad spatial coverage. To our knowledge, this is the first work to predict semantic occupancy forecasts directly from images. Second, we establish a new benchmark on SemanticKITTI for semantic occupancy forecasting from monocular images, a more challenging setting due to its limited field of view and weaker temporal cues. To provide reference points, we adapt two 2D forecasting modules as baselines within our framework. As the first approach to tackle this vision-based task, we propose a novel architecture with a forecasting module that learns future-aware image-plane representations aligned with future observations. These representations are generated through temporal cross-attention across past and current frames, yielding feature maps at the target horizons. The features are first lifted from 2D to 3D using a view transformer, and then refined with a 3D encoder and a feature pyramid neck. Finally, a semantic occupancy head outputs voxel-level semantic categories across multiple horizons. We conduct extensive evaluations of \net{} on both multi-view and monocular settings, present detailed ablation studies to validate the architecture design, and demonstrate consistent improvements over the baselines.

\section{Related Work}\label{sec:related-work}

{\parskip=0pt
\noindent\textit{Scene Forecasting}:
Early work on scene forecasting focused on predicting future semantics directly in the image domain. Several methods extrapolate semantic segmentations into the future~\cite{hurtado2024panoptic}. Beyond 2D semantics, panoptic-depth forecasting introduced the joint prediction of panoptic segmentation and depth for future frames, linking semantics with geometry~\cite{hurtado2024panoptic}. While these works demonstrate the value of forecasting in the image plane, they do not capture how the semantic occupancy of the 3D scene evolves over time, which is critical for autonomous driving~\cite{mohan2024progressive}. Different approaches leverage multi-view images to predict the future in the bird’s-eye view (BEV). Fiery introduced instance-level forecasting in BEV~\cite{hu2021fiery}. PowerBEV~\cite{li2023powerbev} and StretchBEV~\cite{akan2022stretchbev} refined this framework with lightweight architectures and improved spatio-temporal modeling. Beverse~\cite{zhang2022beverse} further unified perception and prediction in a single BEV framework. These methods demonstrate the potential of BEV forecasting, but they remain limited to 2D representations and do not reason about the full 3D semantic occupancy of the scene.

\noindent\textit{LiDAR-based Occupancy Forecasting}:
Another line of work approaches occupancy forecasting from sequential LiDAR point clouds~\cite{hindel2025label}. One method treats point cloud forecasting as a proxy for semantic occupancy~\cite{khurana2023point}. Another proposes modeling the traversal of LiDAR beams through the scene as a differentiable operation, which enables self-supervised training without dense ground truth labels~\cite{khurana2022differentiable}. A dynamics-aware model further extends this idea to urban environments by explicitly capturing motion patterns~\cite{toyungyernsub2022dynamics}. Although geometry-rich, these methods rely on LiDAR, which is expensive and less scalable. Cameras, by contrast, are inexpensive, widely available in autonomous systems, and provide denser semantic cues~\cite{mohan2023neural} alongside geometry, especially with multiple views and temporal context. These advantages motivate our framework for vision-based semantic occupancy forecasting.

\noindent\textit{Image-based Occupancy Forecasting}:
Recent methods extend occupancy prediction into the temporal domain, leading to image-based occupancy forecasting. These approaches typically follow an object-centric view of the scene, dividing it into motion-based categories. For example, Cam4DOcc~\cite{ma2024cam4docc} predicts occupancy using two motion classes: general movable objects and general static objects. OccWorld~\cite{zheng2024occworld} and OCCPROPHET~\cite{chen2025occprophet} build on this idea by forecasting occupancy at multiple levels of granularity, from coarse movable/static groups to more fine-grained motion categories. Motion-based forecasting has clear advantages for modeling dynamics, but it does not explicitly capture the semantic understanding of the scene.

\noindent\textit{Semantic Occupancy Forecasting}:
Semantic occupancy forecasting is a task that predicts the semantic categories of future occupied voxels, rather than focusing on motion-based states as in occupancy forecasting. Instead of modeling object persistence or flow~\cite{luz2024amodal}, it yields a full semantic map of the future scene. For instance, UniOcc~\cite{wang2025uniocc} incorporates temporal reasoning to forecast semantic occupancy, while $I^{2}$-World~\cite{liao20252} leverages inter-frame dependencies to refine predictions. However, these methods operate at the occupancy-map level and must be adapted for vision-centric semantic occupancy forecasting by first generating occupancy maps with vision-based prediction networks. This indirect pipeline limits their ability to fully exploit the rich visual context available in images. In contrast, we propose the first vision-based semantic occupancy forecasting framework that directly leverages images to jointly predict future occupancy and semantics, enabling a more complete and context-aware understanding of 3D scene evolution. Such semantically consistent future scene representations are also beneficial beyond autonomous driving, including applications in robotics~\cite{rfer2026efficient}, healthcare~\cite{mohan2023syn, hindel2025dynamic}, and other domains that require anticipatory spatial scene understanding.\looseness=-1

\begin{figure*}
    \includegraphics[width=0.9\linewidth]{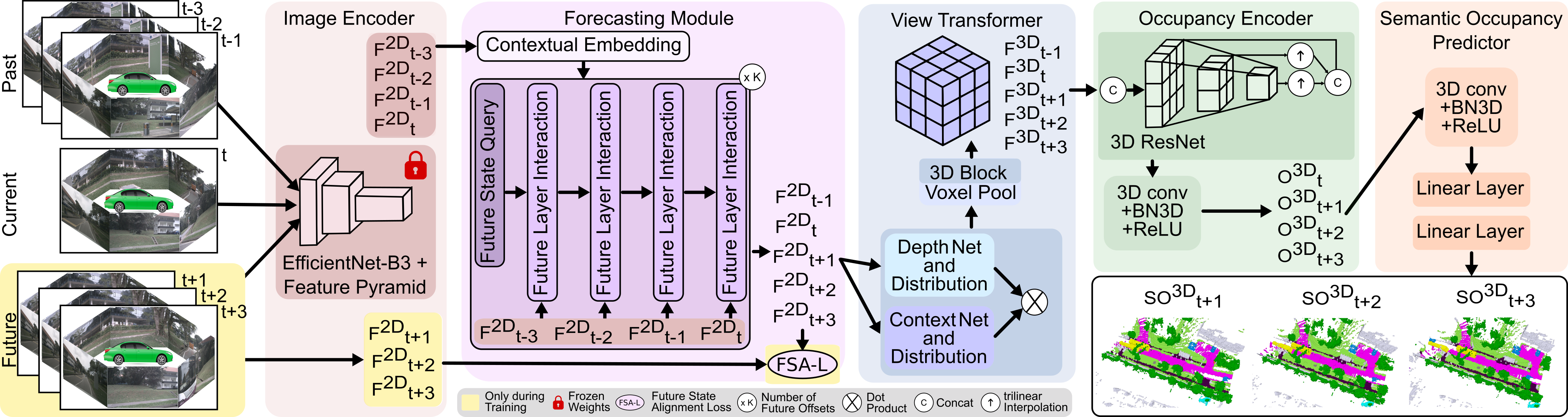}
    \caption{Architecture of \net{} for semantic occupancy forecasting. Multi-view images from the past, current, and future are used as input, with future views only during training. Each image is encoded to 2D features \(F^{2D}\). In the forecasting module, \(F^{2D}\) are embedded as \(\text{Scale}(4,256)\times\text{Views}(M,256)\times\text{Temporal}(4,256)\). The future state query from \(F^{2D}_t\) has shape \((H/16)\times(W/16)\times M\times256\). Each future interaction layer shares weights and contains two multiheaded self-attention blocks, a feedforward network, and a future state synthesizer \((\text{linear}+\text{ReLU})\times2,\ \text{linear}\) with embedding dimension 256 and 16 heads. Depth and context distributions have sizes \(M\times88\times H/16\times W/16\) and \(M\times64\times H/16\times W/16\), yielding a 3D feature volume \(64\times16\times200\times200\). A 3D ResNet occupancy encoder and a semantic predictor output voxelwise logits of size \(N_C\times16\times200\times200\).}
     \label{fig:network}
    \vspace{-0.4cm}
\end{figure*}

\section{\net~Network Architecture }
In this section, we present our \net~architecture, illustrated in~\figref{fig:network}. Our approach introduces a novel forecasting module that learns future image-plane features aligned with future observations. This supports occupancy forecasts while remaining compatible with standard vision-to-3D occupancy prediction pipelines. Here, we describe its implementation within the BEVDet4D framework~\cite{huang2022bevdet4d}, referred to as \textit{BevOcc}. The network encodes synchronized multi-camera inputs through an image encoder, then a forecasting module generates future-aware features at target horizons. A view transformer lifts both current and forecasted features into a 3D voxel grid, which is refined by an occupancy encoder. Finally, a semantic occupancy head outputs voxel-level semantic forecasts for the desired horizons.

\subsection{Problem Formulation}

We address the task of vision-based 3D semantic occupancy forecasting. The input is a sequence of synchronized multi-view images ${I_{i,t} \in \mathbb{R}^{H_i \times W_i \times 3}}$ collected from $M$ cameras over the most recent $N$ time steps ${T-N+1, \ldots, T}$, where $T$ denotes the current time step. Here, $(H_i, W_i)$ represent the image height and width for camera $i$. We assume that camera intrinsic and extrinsic parameters are available for coordinate transformations. In the monocular case, we set $M=1$.

The objective of the task is to predict the semantic occupancy of 3D voxels surrounding the ego-vehicle at future time steps $T+k$, with $k$ representing a prediction horizon in the set $K$. The output is a set of voxel grids $\{O_{T+k} \in \mathbb{R}^{C_{\text{cls}} \times Z \times Y \times X} \mid k \in K\}$, where $C_{\text{cls}}$ is the total number of semantic categories. These include all known object categories (e.g., car, pedestrian, road) along with an additional class for free space, which denotes regions explicitly observed as empty. Additionally, $X$ and $Y$ denote the horizontal spatial dimensions, and $Z$ is the vertical dimension of the voxel grid. In this work, we use $N=4$ input frames as context and generate predictions for future horizons of 1, 2, and 3 seconds.

\subsection{Image Encoder}
The image encoder processes synchronized perspective view inputs from $M$ cameras over the latest $N$ time steps. Each input image is passed through a backbone network followed by a feature pyramid neck to extract multi-scale representations. The backbone generates feature maps at four spatial resolutions $\left(\tfrac{1}{4}, \tfrac{1}{8}, \tfrac{1}{16}, \tfrac{1}{32}\right)$ of the input size. These are fused by the neck into a single feature map at $\tfrac{1}{16}$ resolution. The neck aligns all scales to the common $\tfrac{1}{16}$ resolution by transforming their spatial resolution through convolutional or deconvolutional layers, followed by normalization and ReLU activation, and then concatenates the results along the channel dimension. In \net, we employ EfficientNet-B3~\cite{tan2019efficientnet} as the backbone, which generates feature maps with channel dimensions of $32, 48, 136,$ and $1536$ at the four scales. The feature pyramid neck maps these to $(16, 24, 64, 152)$ channels at $\tfrac{1}{16}$ resolution. Specifically, downsampling is performed using convolutions with kernel sizes $(4, 2, 1)$ and corresponding strides $(4, 2, 1)$ for the $\tfrac{1}{4}, \tfrac{1}{8}, \tfrac{1}{16}$ features, while the $\tfrac{1}{32}$ feature is upsampled using a deconvolution with kernel size and stride $2$. The fused output, obtained by concatenating all scales, generates a feature map with $256$ channels. The resulting temporal multi-camera encoder output is denoted as $\mathbf{F}^{2\mathrm{D}}_{T-N+1:T} \in \mathbb{R}^{N \times M \times C \times H/16 \times W/16}$,  
where $C=256$, and $(H, W)$ denote the height and width of the input image. During training, the image encoder is frozen and its weights are initialized from the pre-trained semantic occupancy prediction network underlying our framework.

\subsection{Forecasting Module}

We propose a forecasting module that synthesizes future representations
${\mathbf{F}^{2\mathrm{D}}{T+k} \mid k \in K}$
directly from a sequence of past and current features
${\mathbf{F}^{2\mathrm{D}}{t} \mid t \in {T-N+1, \ldots, T}}$.
\figref{fig:network} illustrates the overall architecture.
The module is composed of three main stages.
First, the input features are enriched with \textit{Contextual Embeddings} that encode scale, camera view, and temporal position.
Second, a \textit{Future State Synthesizer} integrates temporal information from past to present frames to generate future-aware features for each prediction horizon $k \in K$.
Finally, during training, the \textit{\adloss~Loss} explicitly aligns the synthesized features to match observed future features.
We describe these components in detail in the following subsections.

\paragraph{Contextual Embedding}
The features $\{\mathbf{F}^{2\mathrm{D}}_{t} \mid t \in \{T-N+1, \ldots, T\}\}$ from the image encoder contain information across multiple scales, camera views, and time steps. Without explicit cues, the future state synthesizer cannot effectively distinguish its context. To address this, we augment the features with three sets of learnable embeddings: (i) scale embeddings $\mathbf{E}_{\text{scale}} \in \mathbb{R}^{4 \times 256}$ for the four spatial resolutions, (ii) camera embeddings $\mathbf{E}_{\text{cam}} \in \mathbb{R}^{M \times 256}$ for the $M$ camera views, and (iii) temporal embeddings $\mathbf{E}_{\text{time}} \in \mathbb{R}^{N \times 256}$ for the $N$ past and current frames. Each embedding is broadcast and reshaped to match the spatial layout of the feature maps before being added elementwise. This yields contextually enriched features $\{\mathbf{FE}^{2\mathrm{D}}_{t} \mid t \in \{T-N+1, \ldots, T\}\}$ that explicitly encode the origin of each feature in both temporal and spatial context.

\paragraph{Future State Synthesizer}
The contextually enriched features $\{\mathbf{FE}^{2\mathrm{D}}_{t} \mid t \in \{T-N+1, \ldots, T\}\}$ are used by the future state synthesizer to iteratively construct the latent state of future features $\mathbf{F}^{2\mathrm{D}}_{T+k}$. We initialize forecasting with future state queries, obtained by flattening the spatial dimensions of the current-frame features $\mathbf{FE}^{2\mathrm{D}}_{T}$ into shape $(H/16 \cdot W/16) \times M \times 256$. These queries serve as the initial latent representation and are progressively updated through interactions with the enriched features from past and current frames. This interaction proceeds sequentially, with the queries interacting first with the oldest frame ($T-N+1$) and continuing through to the current frame ($T$). Each interaction step incorporates temporal cues from a specific frame, enabling the model to incrementally build a representation that anticipates the future state.

At each time step, the update is performed by a stack of $L$ \emph{future interaction layers}.
Each layer uses an embedding dimension of 256 and 16 attention heads, and it consists of four key components:
\begin{itemize}
\item A multi-head cross-attention module that enables the evolving future state queries to attend to features from the given time step.
\item A multi-head self-attention module that captures dependencies within the future state queries themselves.
\item A feed-forward network, a two-layer fully connected network with hidden dimensions of 1024 and 256, followed by a ReLU activation.
\item A shared future state synthesizer layer, a three-layer fully connected network with 256 hidden units and ReLU activations. This layer is shared across all $L$ layers within each stack and all $N$ temporal frames.
\end{itemize}
We reshape the updated queries to obtain the final future feature representation $\mathbf{F}^{2\mathrm{D}}_{T+k} \in \mathbb{R}^{M \times 256 \times H/16 \times W/16}$.

\paragraph{\adloss~Loss}
During training, we explicitly align the synthesized future features, $\mathbf{F}^{2\mathrm{D}}_{T+k}$, with their observed counterparts, $\mathbf{G}^{2\mathrm{D}}_{T+k}$. These observed features are obtained by encoding the actual future images with the frozen image encoder. To achieve this, we define an alignment loss $\mathcal{L}_{\text{\short}}$, consisting of a Huber loss $\mathcal{L}_{\text{huber}}$ to enforce magnitude accuracy and a cosine similarity loss $\mathcal{L}_{\text{cosine}}$ to enforce directional alignment in feature space.
The Huber loss is then given by:
\begin{equation}
\mathcal{L}_{\text{huber}} = 
\frac{1}{|K|} \sum_{k \in K}
\begin{cases}
\frac{1}{2} \Delta_{T+k}^2, 
& \text{if } \Delta_{T+k} < \delta, \\[8pt]
\delta \left( \Delta_{T+k} - \tfrac{\delta}{2} \right), 
& \text{otherwise}.
\end{cases},
\end{equation}
where $\Delta_{T+k} = \left\| \mathbf{F}^{2\mathrm{D}}_{T+k} - \mathbf{G}^{2\mathrm{D}}_{T+k} \right\|_2$ and $\delta$ is the Huber threshold.

The cosine similarity loss, $\mathcal{L}_{\text{cosine}}$, is defined as:
\begin{equation}
\mathcal{L}_{\text{cosine}} = 
\frac{1}{|K|} \sum_{k \in K} 
\Bigg( 1 - 
\frac{\mathbf{F}^{2\mathrm{D}}_{T+k} \cdot \mathbf{G}^{2\mathrm{D}}_{T+k}}
{\|\mathbf{F}^{2\mathrm{D}}_{T+k}\|_2 \, \|\mathbf{G}^{2\mathrm{D}}_{T+k}\|_2} \Bigg).
\end{equation}

The overall alignment loss is then defined as:
\begin{equation}
\mathcal{L}_{\text{\short}} = \mathcal{L}_{\text{huber}} + \mathcal{L}_{\text{cosine}},
\end{equation}
which guides the forecasting module to produce future features consistent with observed representations.

\subsection{View Transformer}

The view transformer projects perspective view features into a 3D voxel grid, following the Lift-Splat-Shoot paradigm~\cite{philion2020lift}. At each time step $t$, the 2D features $\mathbf{F}^{2\mathrm{D}}$ are processed by two subnetworks: a context network and a depth network, as in~\cite{huang2022bevdet4d}. The context network generates feature maps $\mathbf{F}^{\text{ctx}} \in \mathbb{R}^{M \times 64 \times H \times W}$, while the depth network predicts a distribution $\mathbf{D} \in \mathbb{R}^{M \times D \times H \times W}$ over $D$ discrete depth bins. Depth-aware features are obtained by element-wise multiplication of $\mathbf{F}^{\text{ctx}}$ with $\mathbf{D}$ and are then unprojected into 3D space using the camera intrinsics and extrinsics. These features are subsequently aggregated into a voxel feature volume $\mathbf{F}^{3\mathrm{D}} \in \mathbb{R}^{64 \times Z \times Y \times X }$ using voxel pooling~\cite{huang2022bevdet4d}, where $(X, Y, Z)$ define the voxel grid resolution. The final output of this module is the set $\{\mathbf{F}^{3\mathrm{D}}_{t} \mid t \in \{T-1, T, T+k, \, k \in K\}\}$. For training, the depth network is supervised by a binary cross-entropy loss ($\mathcal{L}_{\text{depth}}$) between the predicted and ground truth depth distributions.

\subsection{Occupancy Encoder}
The occupancy encoder transforms the 3D voxel features from the view transformer into a richer spatio-temporal representation. The encoder is based on a 3D ResNet-style backbone, composed of 3D bottleneck blocks adapted from the original ResNet architecture~\cite{he2016deep}, where 2D convolution and normalization layers are replaced by their 3D counterparts. 
For each time step, the input feature volume $\mathbf{F}^{3\mathrm{D}}_{t} \in \mathbb{R}^{C \times Z \times Y \times X}$ is first processed by a single 3D bottleneck block, which expands the channel dimension to 64. These processed features from consecutive time steps are then fused to enrich the temporal context. Specifically, the features at time $T$ are concatenated with those from the previous time step $T-1$, resulting in a 128-channel volume $\mathbf{F}^{3\mathrm{D}}_{T} \in \mathbb{R}^{128 \times Z \times Y \times X}$. Similarly, for each future time step $T+k$, the features $\mathbf{F}^{3\mathrm{D}}_{T+k}$ are concatenated with the features from time $T$, producing a 128-channel volume for each future horizon.\looseness=-1

The fused features are then processed by the main 3D backbone, which consists of three stages. The first stage employs a single bottleneck block with 64 channels. The subsequent two stages each comprise two bottleneck blocks. In the second stage, the first block downsamples the spatial resolution by a factor of 2 in each dimension $(X, Y, Z)$ and expands the channel dimension to 128. This is followed by a second block at the same resolution. The third stage follows an identical structure, further downsampling the features and expanding the channels to 256.
To capture multi-scale information, we integrate a 3D feature pyramid neck. The outputs from the downsampling stages are trilinearly upsampled to the base resolution. These upsampled features are then concatenated and fused via a $3\times3\times3$ convolution, followed by BatchNorm3D and ReLU activation. The final output is the encoded occupancy feature volume $\{\mathbf{F}^{3\mathrm{D}}_{t} \mid t \in \{T, T+k,\, k \in K\}\}$, which is subsequently used by the semantic occupancy head.

\subsection{Semantic Occupancy Head}

The semantic occupancy head predicts voxel-level semantics from the output of the occupancy encoder. This is achieved by transforming the input features through a 3D convolutional block and a subsequent lightweight multilayer perceptron (MLP). The 3D convolutional block refines the input features using a 3D convolution layer with a $3 \times 3 \times 3$ kernel, followed by BatchNorm3D and a ReLU activation. This block outputs a 64-channel feature volume. The refined per-voxel features are then passed to a lightweight MLP composed of a fully connected layer with 128 hidden units and a Softplus activation, followed by a final fully connected layer that outputs $C_{\text{cls}}$ classes. The final prediction output is denoted as $\mathbf{SO} \in \mathbb{R}^{C_{\text{cls}} \times Z \times Y \times X}$. The head is applied to the 3D feature volumes at both the current frame $T$ and the forecasted horizons $T+k$ for $k \in K$. The task-level training objective is defined as a voxel-wise cross-entropy loss:
\begin{equation}
\mathcal{L}_{\text{task}} = \mathrm{CE}(\mathbf{SO}_{t}, \mathbf{SO}^{\star}_{t}) + \sum_{k \in K} \mathrm{CE}(\mathbf{SO}_{t+k}, \mathbf{SO}^{\star}_{t+k}),
\end{equation}
where $\mathbf{SO}^{\star}$ denotes the ground truth semantic occupancy.

\section{Experimental Evaluation}
In this section, we first describe the experimental setup, including datasets, baselines, and training protocols. We then present quantitative comparisons and benchmarking results, followed by detailed ablation studies on the proposed architectural components. Finally, we provide qualitative comparisons.

\subsection{Experimental Setup}
\paragraph{Datasets} We evaluate our approach on two widely used autonomous driving datasets: Occ3D-nuScenes~\cite{tian2023occ3d} and SemanticKITTI~\cite{behley2019semantickitti}. Occ3D-nuScenes provides 700 training scenes and 150 validation scenes. Each frame contains six images from multi-view cameras and a 32-beam LiDAR point cloud. The dataset includes dense voxel-wise semantic occupancy annotations. These annotations are defined on a $200 \times 200 \times 16$ voxel grid. Each voxel measures $\SI{0.4}{\meter} \times \SI{0.4}{\meter} \times \SI{0.4}{\meter}$. The semantic labels cover 17 categories, including 16 object classes and one free-space class. In comparison, SemanticKITTI provides LiDAR point clouds with synchronized front-view camera images. It is split into 10 training sequences and 1 validation sequence. The ground-truth semantic occupancy is defined on voxel grids of size $256 \times 256 \times 32$. This dataset has a finer voxel size of $\SI{0.2}{\meter} \times \SI{0.2}{\meter} \times \SI{0.2}{\meter}$. It contains 20 categories in total, with 19 semantic classes and one free-space class.

\paragraph{Baselines} We formulate two vision-based semantic occupancy forecasting baselines to benchmark against our proposed \net{}, referred to as \textit{Naive} and PDCast. For both, we retain the same semantic occupancy prediction network as \net{} and employ \adloss{} to align future image-plane features with their corresponding observations during training, replacing only our forecasting module. The \textit{Naive} baseline performs spatiotemporal feature aggregation and iterative forecasting: encoder features $\mathbf{F}^{2\mathrm{D}}_{T-N+1:T}$ are projected with $1\times1$ convolutions, concatenated, and fused with two $3\times3$ Conv-BN-ReLU blocks. These fused features are then combined with $\mathbf{F}^{2\mathrm{D}}_{T}$ via a $1\times1$ deprojection to yield the first forecast, and subsequent steps are generated by reprojecting and refining the previous prediction with the same convolution blocks. The PDCast baseline directly replaces our forecasting module with the one from~\cite{hurtado2024panoptic}. Finally, to demonstrate the transferability of our forecasting module to other vision-to-3D occupancy pipelines, we replace the BEVOcc backbone with COTR~\cite{ma2024cotr} by inserting our forecasting module after its featurizer, which we denote as \net{}-COTR.

\paragraph{Training Protocol}
We initialize the weights of \net{} from the pretrained BevOcc semantic occupancy prediction network. During training, the image encoder is kept frozen. We use the AdamW optimizer with an initial learning rate of $1\mathrm{e}{-3}$ for the forecasting module and $1\mathrm{e}{-5}$ for the remaining layers, and train for $24$ epochs. After $12$ epochs, the learning rate of the forecasting module is reduced to $1\mathrm{e}{-4}$. The input image resolution is set to $256\times704$ for Occ3D-nuScenes and $192\times640$ for SemanticKITTI. We set $M=6$ for Occ3D-Nuscenes and $M=1$ for SemanticKITTI, and fix the Huber threshold at $\delta=2$. The network is trained with a batch size of $4$ and the total loss
\begin{equation}
\mathcal{L} = \mathcal{L}_{\text{task}} + \alpha \, \mathcal{L}_{\text{FSA}}, \quad \alpha = 30.
\end{equation}

For pretraining BevOcc, we follow dataset-specific protocols. On Occ3D-nuScenes, BevOcc is trained following the COTR~\cite{ma2024cotr} setup, using AdamW with an initial learning rate of $1\mathrm{e}{-4}$, weight decay of $0.01$, and a batch size of $8$. On SemanticKITTI, BevOcc is trained following the OccFormer~\cite{zhang2023occformer} setup, with the same optimizer and weight decay but a batch size of $4$.

\begin{table*}
    \setlength{\tabcolsep}{0.012\linewidth}
    \centering
    \caption{Semantic occupancy forecasting results on the Occ3D-nuScenes validation set. We report mIoU and IoU at 1s, 2s, and 3s horizons, with Avg. denoting the mean. Input Type refers to Occupancy-input (pre-computed occupancy maps) and Image-input (multi-view images). Base Occupancy indicates the underlying predictor: STC~\cite{liao2025stcocc}, BevOcc (COTR~\cite{ma2024cotr} setup), or COTR. \net{} (ours) consistently outperforms both input types.}
    \begin{tabular}{l|c|c|cccc|cccc|c}
        \toprule
        \multirow{2}{*}{Method} & \multirow{2}{*}{Input Type} & \multirow{2}{*}{Base Occupancy} & \multicolumn{4}{c|}{mIoU (\%)} &  \multicolumn{4}{c|}{IoU (\%)} & FPS \\
        & & & 1s & 2s & 3s & Avg. & 1s & 2s & 3s & Avg. \\
        \midrule
        Naive                     & Image-input & BevOcc & 13.23 & 9.21 & 7.89 & 10.11 & 23.75 & 20.92 & 18.65 & 21.10 & 7.3\\
        OccWorld~\cite{zheng2024occworld}                & Occupancy-input & STC   & 10.97 & 15.16 & 10.19 & 12.10  & 20.68 & 24.81 & 20.09 & 21.86 & 3.77 \\
        DOME~\cite{gu2024dome}                      & Occupancy-input & STC   & 17.79 & 14.23 & 11.58 & 14.53 & 26.39 & 23.20 & 20.42 & 23.33 & 2.75\\
        PDCast~\cite{hurtado2024panoptic}                    & Image-input & BevOcc & 18.36 & 14.91 & 12.74 & 15.33 & 30.52 & 28.84 & 26.12 & 28.49 & 3.9\\
        $I^{2}$-World~\cite{liao20252}             & Occupancy-input & BevOcc   & 20.54 & 17.67 & 15.42 & 17.87 & 29.72 & 27.49 & 26.10 & 27.77 & 4.73 \\
        $I^{2}$-World~\cite{liao20252}             & Occupancy-input & STC   & 21.67 & 18.78 & 16.47 & 18.97 & 32.66 & 28.76 & 26.99 & 28.77 & 4.21 \\
        \net{} (Ours)          & Image-input & BevOcc & 22.71 & 19.34 & 17.01 & 19.68 & 34.64 & 32.23 & 30.41 & 32.42 & 6.05\\
        \net{} (Ours)          & Image-input & COTR   & \textbf{24.86} & \textbf{20.73} & \textbf{17.88} & \textbf{21.15} & \textbf{35.25} & \textbf{33.53} & \textbf{31.36} & \textbf{33.38} & 1.34\\
        \bottomrule
    \end{tabular}
    \label{tab:mainresults_nuforecast}
    \vspace{-0.3cm}
\end{table*}

\begin{table}
    \setlength{\tabcolsep}{0.005\linewidth}
    \centering
    \caption{Semantic occupancy forecasting results on the SemanticKITTI validation set. The table presents our newly introduced benchmark for forecasting from monocular camera images, reporting mIoU and IoU at 1s, 2s, and 3s horizons, with Avg. denoting the mean over all horizons.}
    \begin{tabular}{l|cccc|cccc}
        \toprule
        \multirow{2}{*}{Method}  & \multicolumn{4}{c|}{mIoU (\%)} &  \multicolumn{4}{c}{IoU (\%)} \\
        &  1s & 2s & 3s & Avg. & 1s & 2s & 3s & Avg. \\
        \midrule
        Naive                 & 5.32 & 4.12 &  2.69 & 4.04 & 25.31 & 21.23 & 19.45 & 21.99   \\
        PDCast~\cite{hurtado2024panoptic}               &     7.48 & 5.72  & 4.58 & 5.92 & 32.44 & 29.51 & 27.65 & 29.86 \\
        \net{} (Ours)          & 9.10 & 7.22 & 6.17 & 7.49 & 33.38 & 31.38 & 29.78 & 31.51 \\
        \net{}-COTR (Ours)         & \textbf{9.89} & \textbf{7.83} & \textbf{6.58} & \textbf{8.10} & \textbf{35.05} & \textbf{32.76} & \textbf{30.42} & \textbf{32.74} \\
        \bottomrule
    \end{tabular}
    \label{tab:mainresults_semforecast}
\end{table}

\begin{table}[t]
\setlength{\tabcolsep}{2pt}
\centering
\caption{Effect of task loss $\mathcal{L}{\text{task}}$ and alignment loss $\mathcal{L}{\text{\short}}$ on forecasting performance, evaluated on Occ3D-nuScenes.}
\begin{tabular}{c c|cccc|cccc}
\toprule
$\mathcal{L}_{\text{task}}$ & $\mathcal{L}_{\text{\short}}$ & \multicolumn{4}{c|}{mIoU} & \multicolumn{4}{c}{IoU} \\
 & & 1s & 2s & 3s & Avg. & 1s & 2s & 3s & Avg. \\
\midrule
\checkmark &  & 13.19 & 11.23 & 9.87 & 11.43 & 24.46 & 22.76 & 21.47 & 22.89   \\
 & \checkmark & 20.81 & 17.72 & 15.58 & 18.03 & 32.20 & 29.95  & 28.26 & 30.14 \\
\checkmark & \checkmark & \textbf{22.71} & \textbf{19.34} & \textbf{17.01} & \textbf{19.68} & \textbf{34.64} & \textbf{32.23} & \textbf{30.41} & \textbf{32.42}  \\
\bottomrule
\label{tab:loss}
\end{tabular}
\vspace{-0.33cm}
\end{table}

\begin{table}[t]
\setlength{\tabcolsep}{2pt}
\centering
\caption{Contribution of individual \adloss{} components to forecasting performance. Results are reported on the Occ3D-nuScenes validation set.}
\begin{tabular}{c c|cccc|cccc}
\toprule
$\mathcal{L}_{\text{huber}}$ & $\mathcal{L}_{\text{cosine}}$ & \multicolumn{4}{c|}{mIoU} & \multicolumn{4}{c}{IoU} \\
 & & 1s & 2s & 3s & Avg. & 1s & 2s & 3s & Avg. \\
\midrule
\checkmark & & 22.13 & 18.84 &  16.57 & 19.18 & 34.37 & 31.98 & 30.17 & 32.18   \\
 & \checkmark & 21.93 & 18.67  & 16.42 & 19.01 & 34.32 & 31.93 & 30.13 & 32.13 \\
\checkmark & \checkmark & \textbf{22.71} & \textbf{19.34} & \textbf{17.01} & \textbf{19.68} & \textbf{34.64} & \textbf{32.23} & \textbf{30.41} & \textbf{32.42}  \\
\bottomrule
\label{tab:fsa}
\end{tabular}
\end{table}

\begin{table}[t]
\setlength{\tabcolsep}{2pt}
\centering
\caption{Comparison of different initialization strategies for future state queries: learned parameters vs. features from the current frame $\mathbf{F}^{2\mathrm{D}}_{T}$. Results are reported on the Occ3D-nuScenes~\cite{tian2023occ3d} validation set.}
\label{tab:forecasthead-init}
\begin{tabular}{l|ccc|c|ccc|c}
\toprule
Variant & \multicolumn{4}{c|}{mIoU} & \multicolumn{4}{c}{IoU} \\
 & 1s & 2s & 3s & Avg. & 1s & 2s & 3s & Avg. \\
\midrule
Learned & 21.18  & 18.03 & 15.86 & 18.36 & 33.89 & 31.53 & 29.75 & 31.72  \\
$\mathbf{F}^{2\mathrm{D}}_{T}$ & \textbf{22.71} & \textbf{19.34} & \textbf{17.01} & \textbf{19.68} & \textbf{34.64} & \textbf{32.23} & \textbf{30.41} & \textbf{32.42}  \\
\bottomrule
\end{tabular}
\vspace{-0.5cm}
\end{table}

\begin{table}[t]
\setlength{\tabcolsep}{2pt}
\centering
\caption{Effect of contextual embeddings on forecasting performance. Results are reported on Occ3D-nuScenes~\cite{tian2023occ3d} validation set.}
\label{tab:forecasthead-emb}
\begin{tabular}{ccc|ccc|c|c}
\toprule
Temporal & Scale & Camera & \multicolumn{4}{c|}{mIoU} & \multicolumn{1}{c}{IoU} \\
 & & & 1s & 2s & 3s & Avg. &  Avg. \\
\midrule
 -- & -- &  --  & 20.67 & 17.60 & 15.48 & 17.91 &  31.49  \\
\checkmark & -- & -- & 21.50 & 18.31 & 16.10 & 18.63 &  31.97  \\
-- &\checkmark & -- &21.63 & 18.42 & 16.20 & 18.75 & 32.04  \\
-- & -- &  \checkmark & 21.89 & 18.64 & 16.39 & 18.97 & 32.12 \\
\checkmark & -- &\checkmark & 22.36 & 19.04 & 16.74 & 19.38 &  32.21  \\
-- &\checkmark &\checkmark & 22.47 & 18.73 & 16.47 & 19.22 & 32.27  \\
\checkmark&\checkmark &\checkmark & \textbf{22.71} & \textbf{19.34} & \textbf{17.01} & \textbf{19.68} & \textbf{32.42} \\
\bottomrule
\end{tabular}
\end{table}

\begin{table}[t]
\setlength{\tabcolsep}{2pt}
\centering
\caption{Impact of the number of future interaction layers $L$ in the forecasting module, evaluated on the Occ3D-nuScenes~\cite{tian2023occ3d}.}
\label{tab:forecasthead-layers}
\begin{tabular}{l|ccc|c|ccc|c}
\toprule
$L$ & \multicolumn{4}{c|}{mIoU} & \multicolumn{4}{c}{IoU} \\
 & 1s & 2s & 3s & Avg. & 1s & 2s & 3s & Avg. \\
\midrule
1 & 19.47  & 16.58 & 14.58 & 16.87 & 31.65 & 29.44 & 27.78 & 29.62 \\
2 & 21.87 & 18.62 & 16.38 & 18.95 & 34.30 & 31.91 & 30.11 & 32.11 \\
3 & \textbf{22.71} & \textbf{19.34} & \textbf{17.01} & \textbf{19.68} & \textbf{34.64} & \textbf{32.23} & \textbf{30.41} & \textbf{32.42}  \\
4 & 22.38 & 19.05 & 16.76 & 19.40 & 34.62 & 32.21 & 30.39 & 32.41 \\
\bottomrule
\end{tabular}
\vspace{-0.5cm}
\end{table}

\subsection{Benchmarking Results}
Tab.~\ref{tab:mainresults_nuforecast} reports vision-based semantic occupancy forecasting results on the Occ3D-nuScenes validation set. We consider two families of methods. The first, which we refer to as \textit{occupancy-input} methods, use a vision-based semantic occupancy predictor (STC~\cite{liao2025stcocc} or BevOcc) to generate occupancy grids that are then forecasted by the respective approaches. The second, referred to as \textit{image-input} methods, directly forecast semantic occupancy from images. For these, we evaluate BevOcc and COTR~\cite{ma2024cotr} as backbone semantic occupancy networks, each incorporated with different forecasting modules employed in the image feature space. We report performance in terms of mean Intersection-over-Union (mIoU) and Intersection-over-Union (IoU) across \SI{1}{\second}, \SI{2}{\second}, and \SI{3}{\second} prediction horizons. Additionally, we report the runtime in frames per second (FPS) for each prediction horizon, measured on an RTX 3090 GPU following~\cite{zheng2024occworld}.

We observe that \net{} consistently outperforms both occupancy-input and image-input baselines across all horizons. Occupancy-input approaches such as OccWorld and DOME perform poorly (12.10\% and 14.53\% mIoU) because their reliance on pre-computed occupancy maps leads to the loss of complementary fine-grained visual and temporal information that would otherwise aid forecasting. I$^{2}$-World mitigates this limitation by introducing a continuous scene tokenizer that better preserves spatial structure and temporal context. This leads to improved performance, reaching 17.87\% mIoU with BevOcc and 18.97\% with STC. In contrast, \net{} leverages its future state synthesizer to exploit these complementary cues directly. By progressively integrating past and current features through cross-attention, it provides stronger temporal reasoning. This design outperforms the Naive baseline, which relies on shallow convolutions, and PDCast, which depends on pre-aggregated temporal features with windowed self-attention. With BevOcc, \net{} achieves 19.68\% mIoU and 32.42\% IoU at 6.05 FPS. Pairing it with COTR~\cite{ma2024cotr} further improves performance to 21.15\% mIoU and 33.38\% IoU, though this comes at the cost of a slower runtime of 1.34 FPS.

We further evaluate our approach on the more stringent SemanticKITTI benchmark in Tab.~\ref{tab:mainresults_semforecast}, which poses a greater challenge due to its monocular front-view input. Unlike multi-view datasets such as Occ3D-nuScenes, this setting provides limited spatial coverage and weaker temporal cues. Despite these constraints, \net{} consistently outperforms Naive and PDCast across all horizons, achieving 7.49\% mIoU and 31.51\% IoU on average. Incorporating COTR as the backbone further improves results to 8.10\% mIoU and 32.74\% IoU. These results show that our forecasting module remains effective under limited visual context, providing a strong foundation for future work in monocular forecasting from a single camera view.

\subsection{Ablation Study}
In this section, we present two ablation experiments on key components of our network.
\subsubsection{Loss Components}
Our proposed \net{} is trained with two supervised losses: the task-level loss $\mathcal{L}_{\text{task}}$ and the \adloss{} loss $\mathcal{L}_{\text{\short}}$. To understand the roles of the two loss functions, we perform an ablation study on their contributions to model performance, as reported in Tab.~\ref{tab:loss}. Using only $\mathcal{L}_{\text{task}}$ yields limited performance of 11.43\% mIoU and 22.89\% IoU on average, indicating that supervision solely from ground truth occupancy at the prediction horizon is insufficient. Since our approach forecasts in image feature space while the downstream task layers are simultaneously being trained, the gradient signal becomes sparse. We speculate that this occurs because the forecasted features must already be sufficiently semantically informative for the downstream heads, which imposes a limiting constraint. In contrast, training with $\mathcal{L}_{\text{\short}}$, where the downstream layers are frozen to minimize task-level supervision, directly optimizes the forecasting module through feature-level alignment with future state observations. This configuration substantially improves performance, boosting the average mIoU to 18.03\% and IoU to 30.14\%. Finally, combining both losses achieves the best results of 19.68\% mIoU and 32.42\% IoU, showing that while $\mathcal{L}_{\text{\short}}$ drives the majority of the learning, the task-level loss provides complementary guidance that further refines long-horizon predictions.\looseness=-1

We further evaluate different variants of the $\mathcal{L}_{\text{\short}}$ in Tab.~\ref{tab:fsa}. Both Huber  and Cosine losses ($\mathcal{L}_{\text{huber}}$, $\mathcal{L}_{\text{cosine}}$) contribute positively, with $\mathcal{L}_{\text{huber}}$ performing slightly better. The best performance of 19.68\% mIoU and 32.42\% IoU is achieved when both are combined, confirming that complementary distance measures provide stronger alignment supervision.

\begin{figure}[t]
\setlength{\tabcolsep}{0.005cm}
\begin{tabular}{P{0.25cm} P{2.78cm} P{2.78cm} P{2.78cm}}
  & \scriptsize $k = 1s$ & \scriptsize $k = 2s$ & \scriptsize $k = 3s$ \\
    \rotatebox{90}{\scriptsize Image} &
    \includegraphics[width=\linewidth]{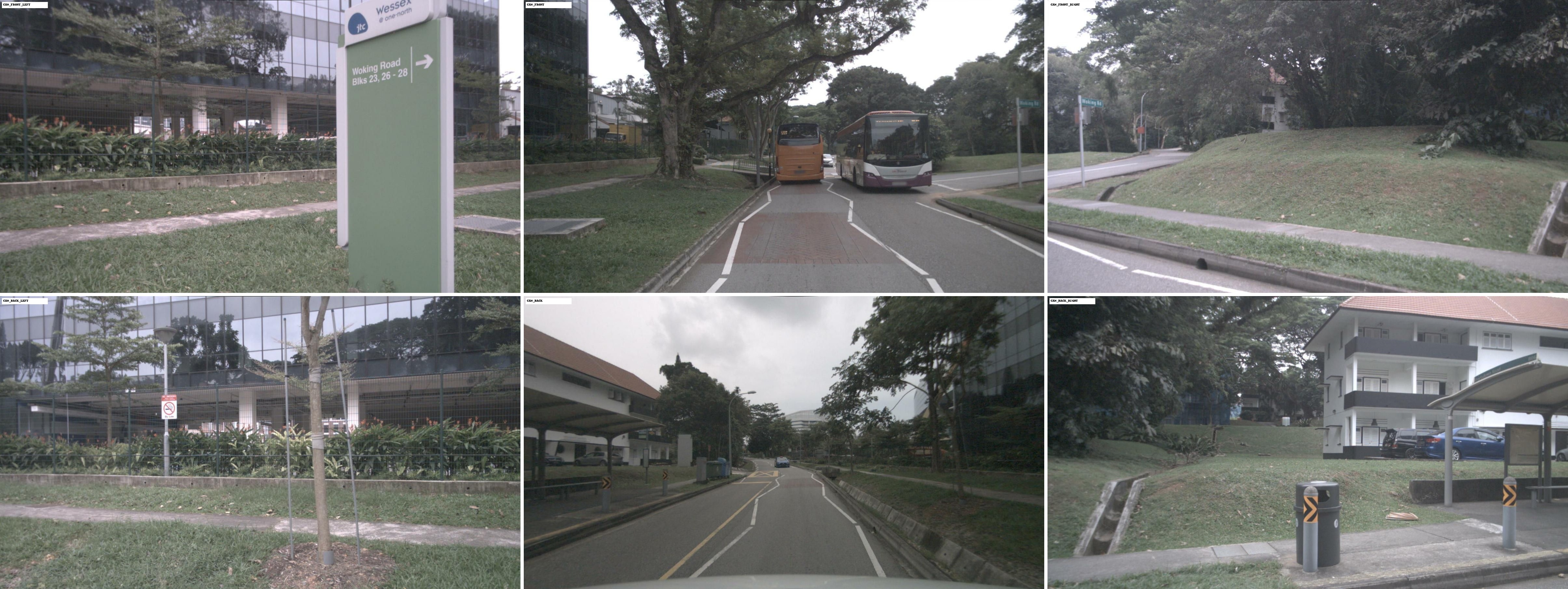} &
    \includegraphics[width=\linewidth]{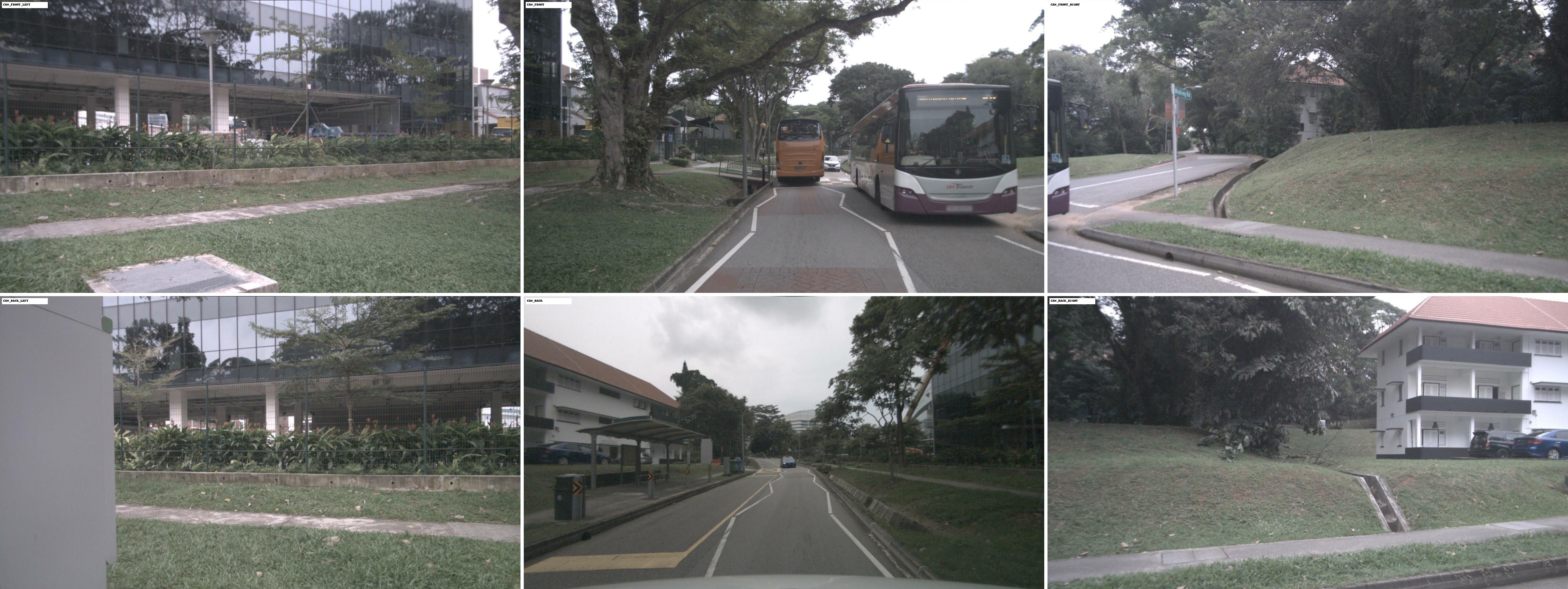} &
    \includegraphics[width=\linewidth]{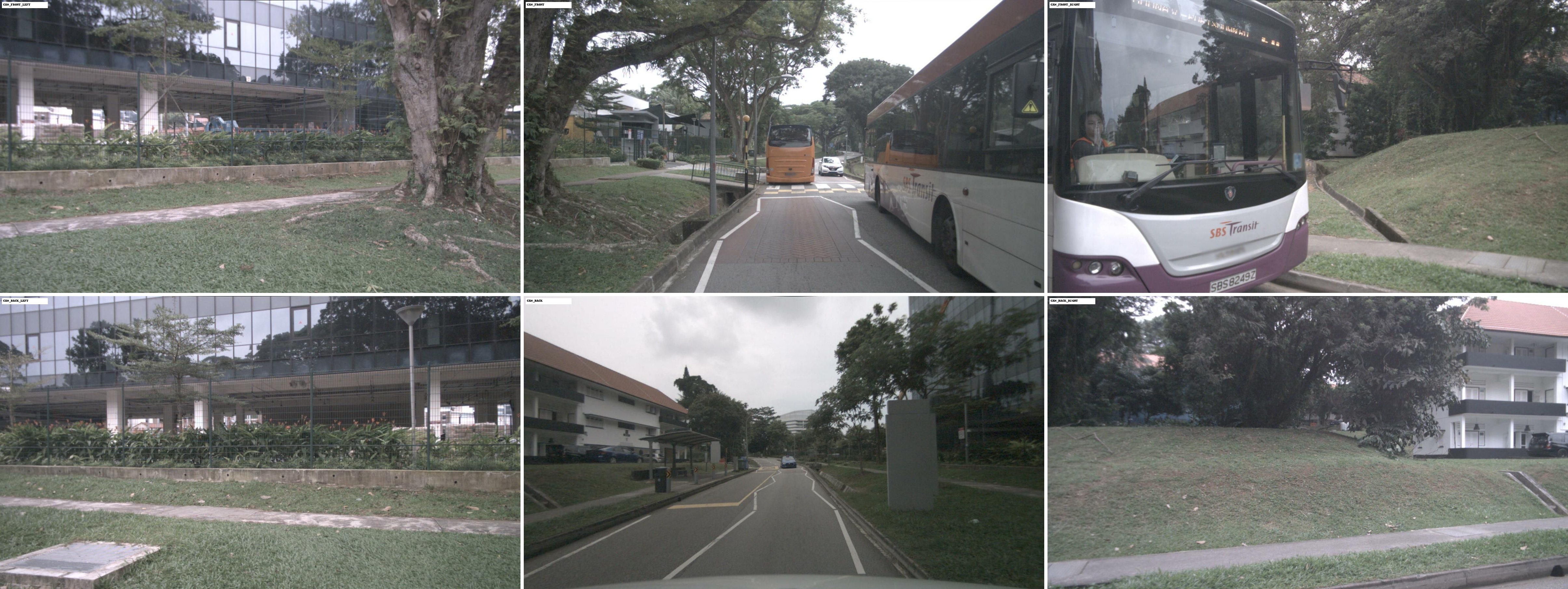} \\
    \rotatebox{90}{\makebox[1.5cm][c]{\scriptsize GT}}  &
    \includegraphics[width=1.0\linewidth, trim=350pt 300pt 300pt 100pt, clip]{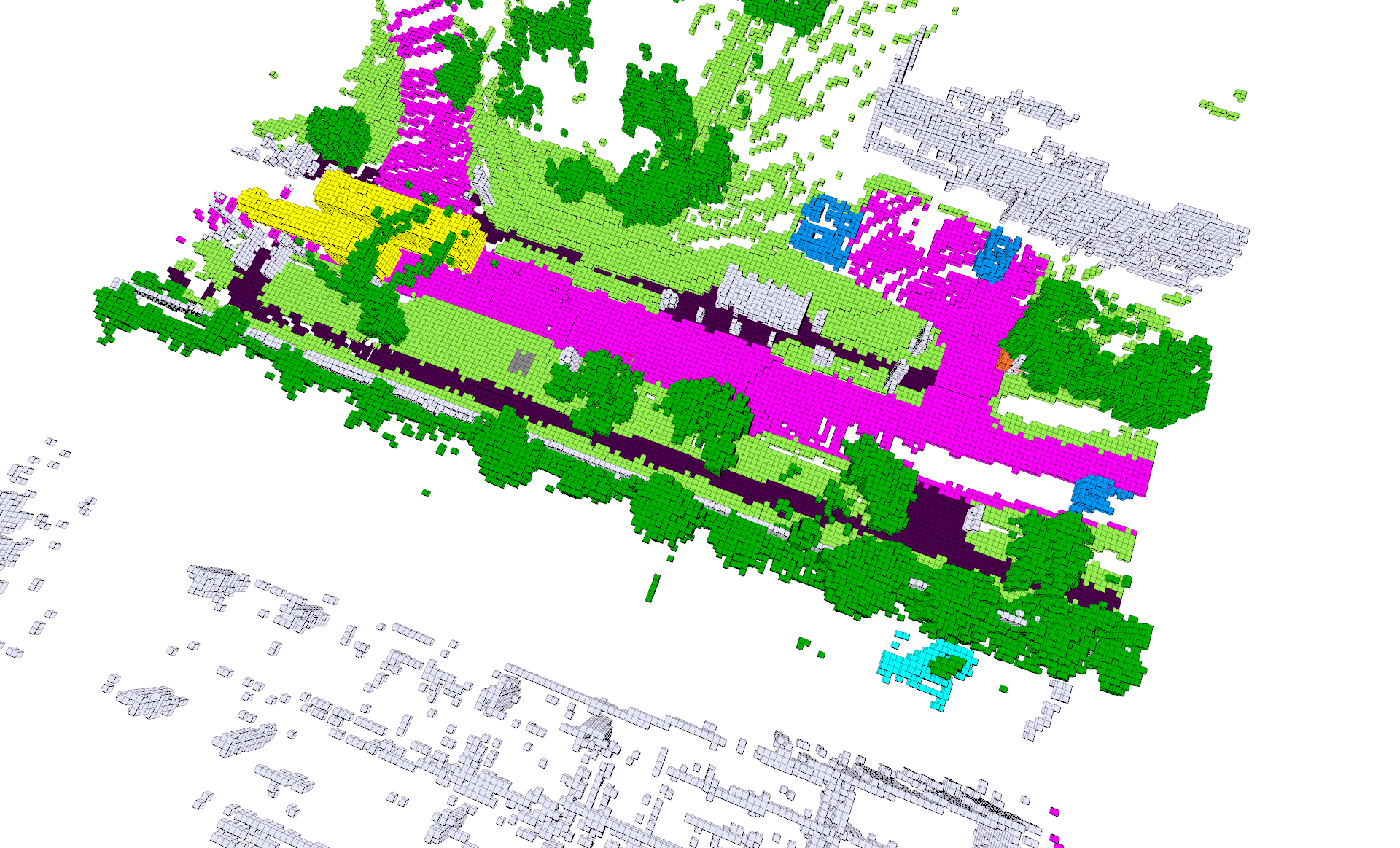}&
    \includegraphics[width=1.0\linewidth, trim=350pt 300pt 300pt 100pt, clip]{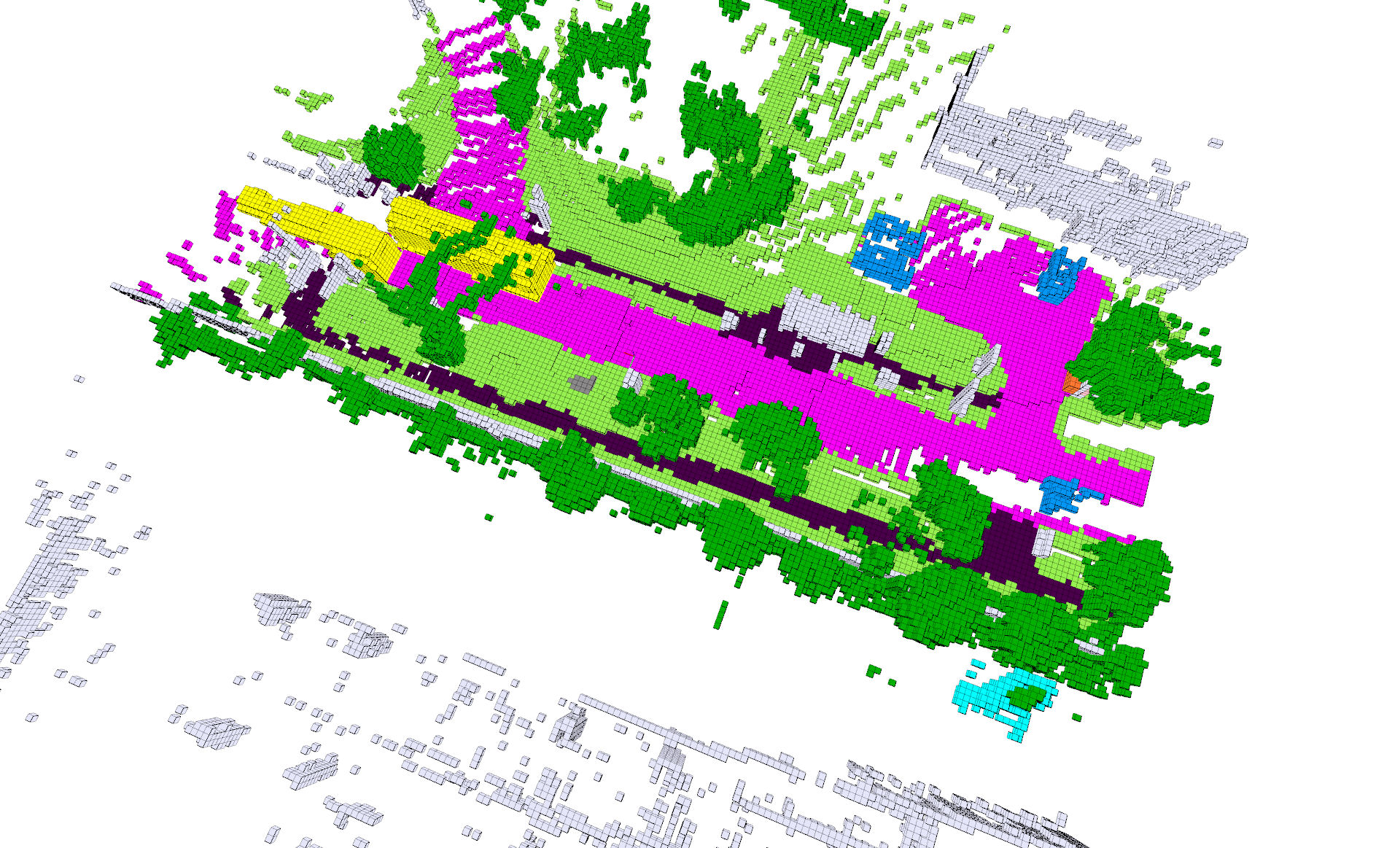} &
    \includegraphics[width=1.0\linewidth, trim=350pt 300pt 300pt 100pt, clip]{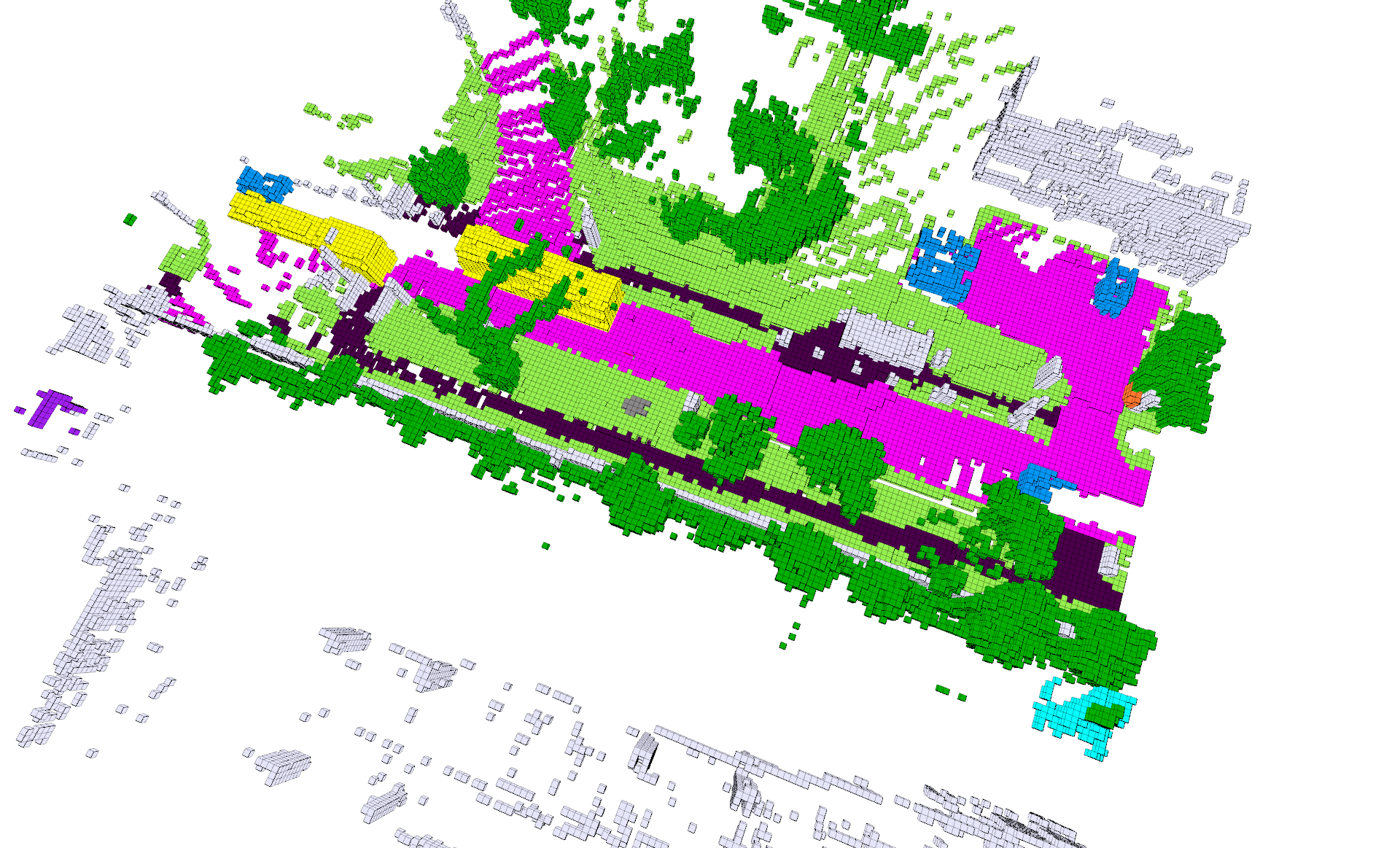}\\

    \rotatebox{90}{\makebox[1.5cm][c]{\scriptsize $I^{2}$-World}}  &
    \includegraphics[width=1.0\linewidth, trim=350pt 300pt 300pt 100pt, clip]{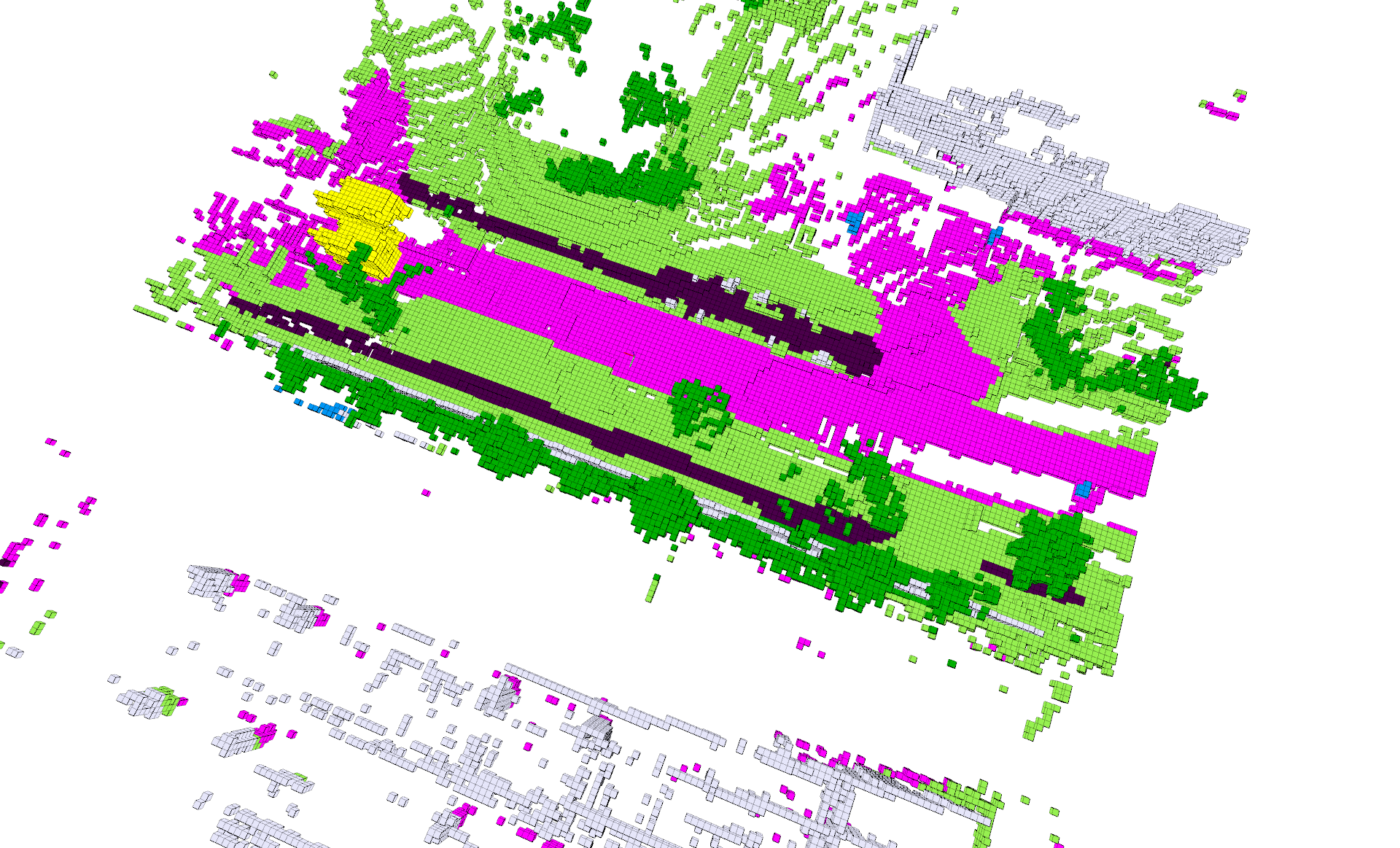}&
    \includegraphics[width=1.0\linewidth, trim=350pt 300pt 300pt 100pt, clip]{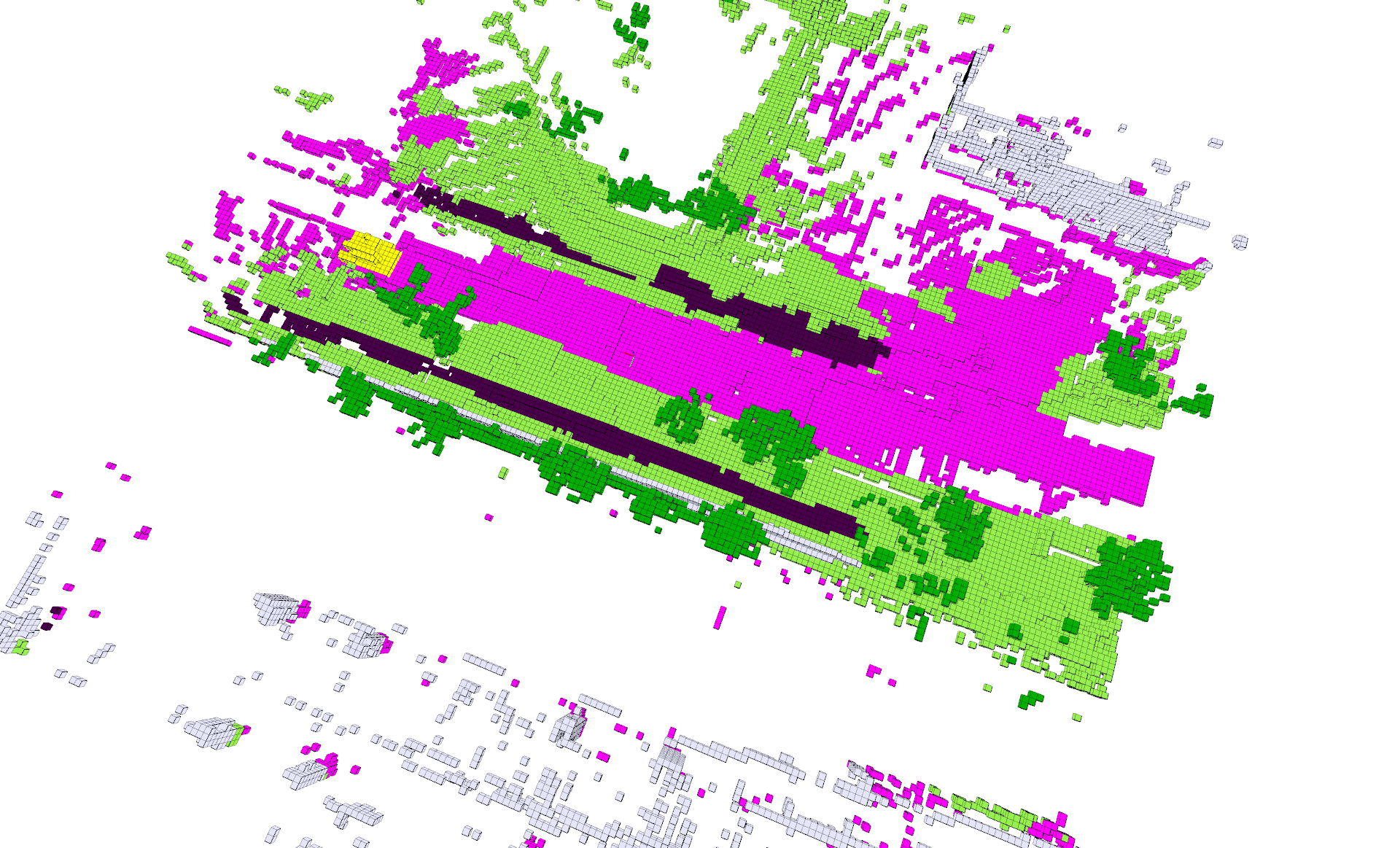} &
    \includegraphics[width=1.0\linewidth, trim=350pt 300pt 300pt 100pt, clip]{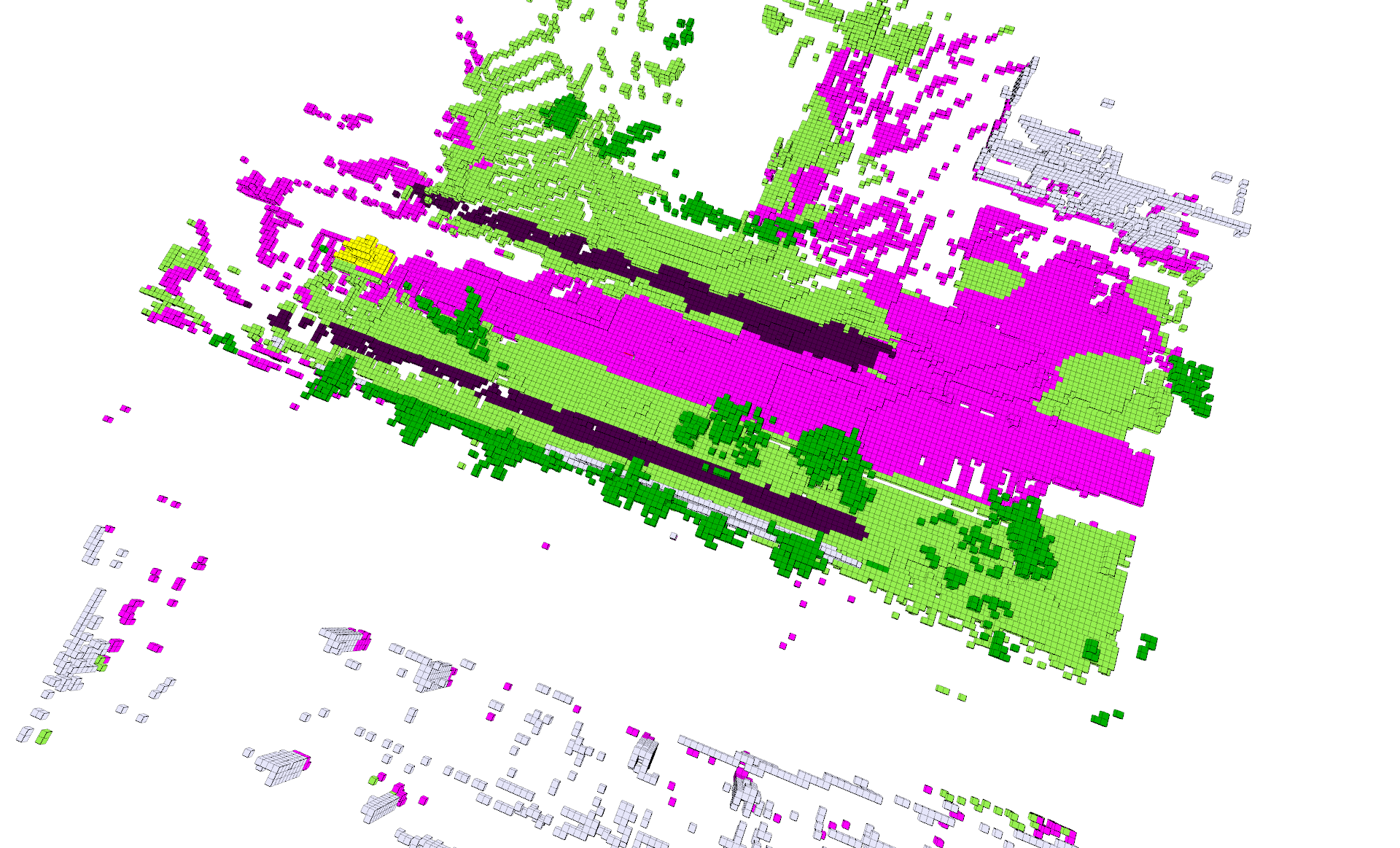}\\
    \rotatebox{90}{\scriptsize \net (ours)} &
    \includegraphics[width=1.0\linewidth, trim=350pt 300pt 300pt 100pt, clip]{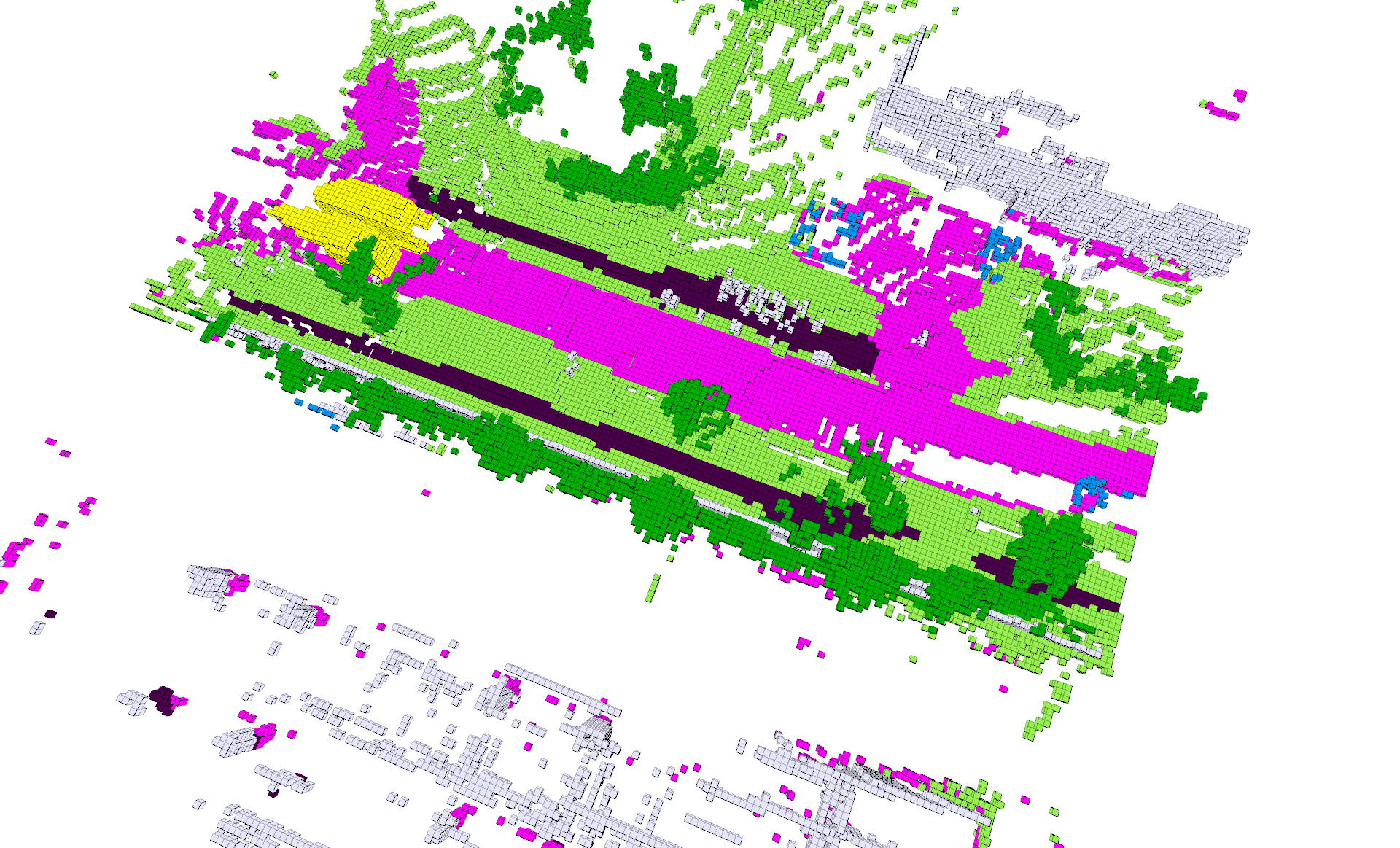}&
    \includegraphics[width=1.0\linewidth, trim=350pt 300pt 300pt 100pt, clip]{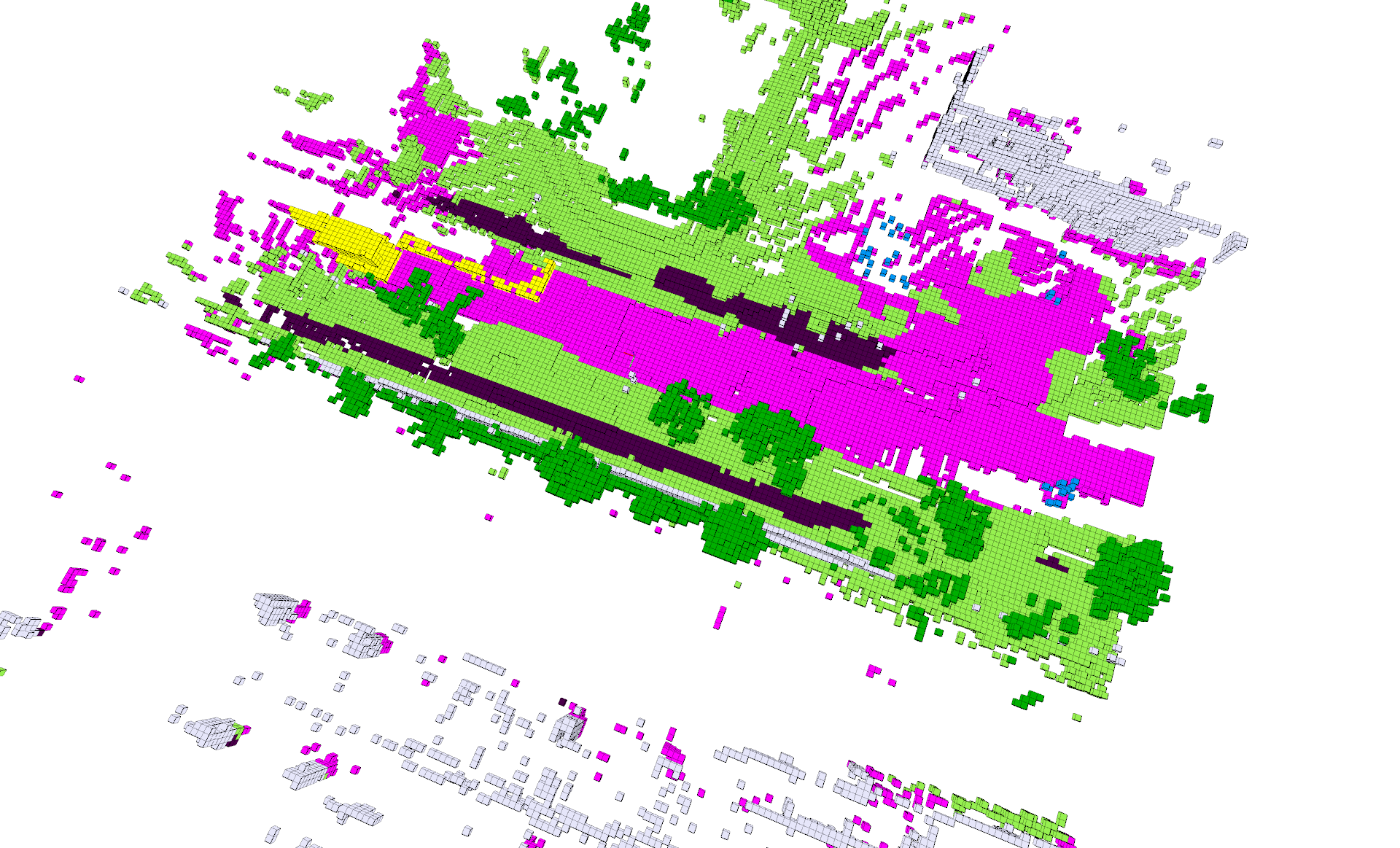} &
    \includegraphics[width=1.0\linewidth, trim=350pt 300pt 300pt 100pt, clip]{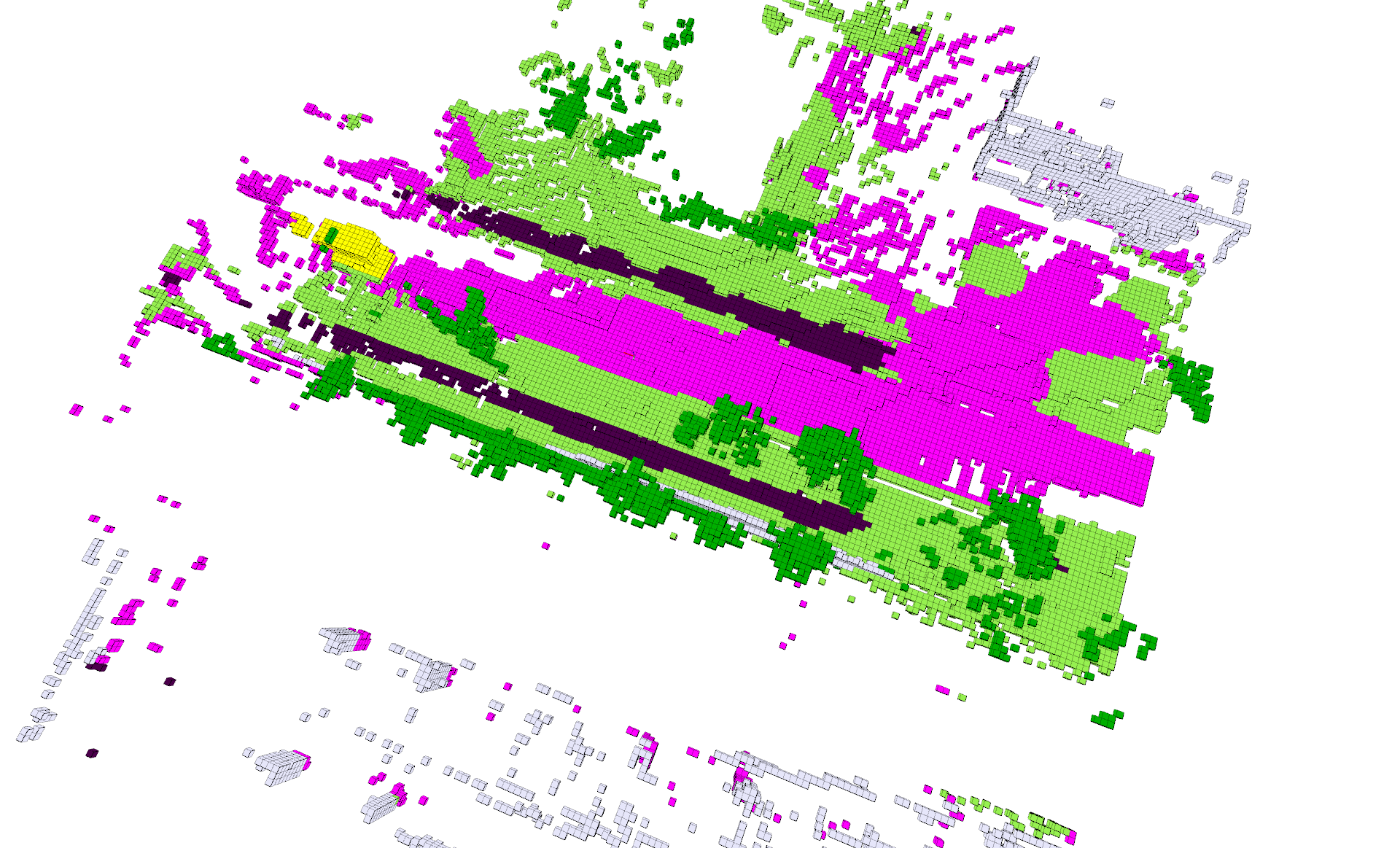}\\
\end{tabular}
\caption{Qualitative comparison of predictions from our proposed \net{} model with the second-best baseline $I^{2}$-World on the  Occ3D-nuScenes dataset. We show the multi-view camera images corresponding to the future frame at $T + k$ and the semantic occupancy ground truth (GT).}
\label{fig:qualitative-nuscenes}
\end{figure}

\begin{figure}
\setlength{\tabcolsep}{0.005cm}
\begin{tabular}{P{0.24cm} P{2.8cm} P{2.8cm} P{2.8cm}}
  & \scriptsize $k = 1s$ & \scriptsize $k = 2s$ & \scriptsize $k = 3s$ \\
    \rotatebox{90}{\scriptsize Image} &
    \includegraphics[width=\linewidth]{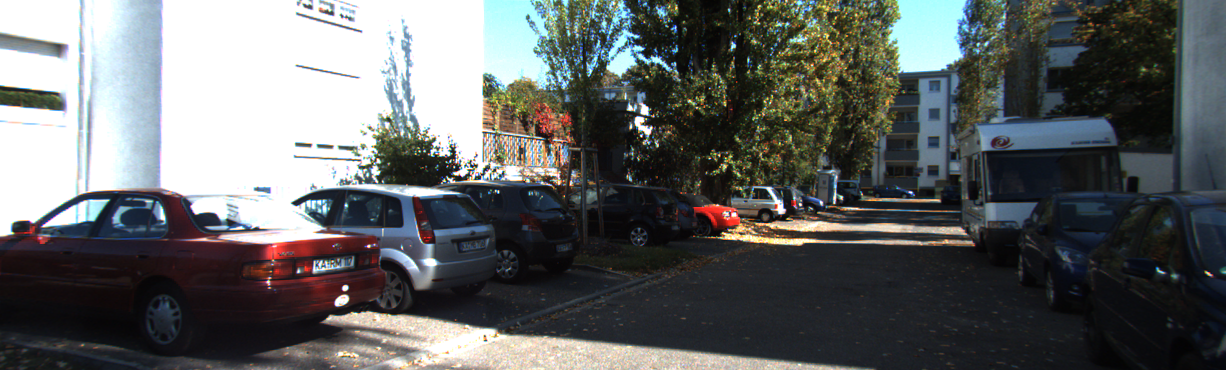} &
    \includegraphics[width=\linewidth]{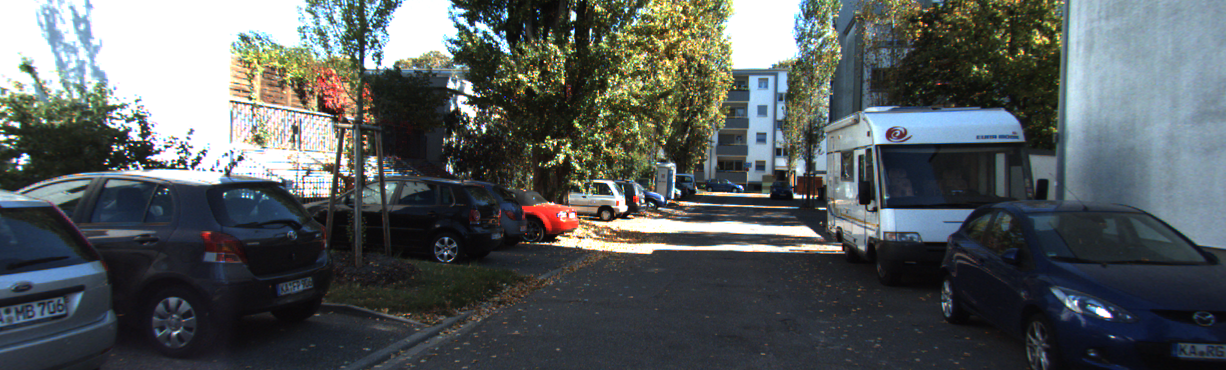} &
    \includegraphics[width=\linewidth]{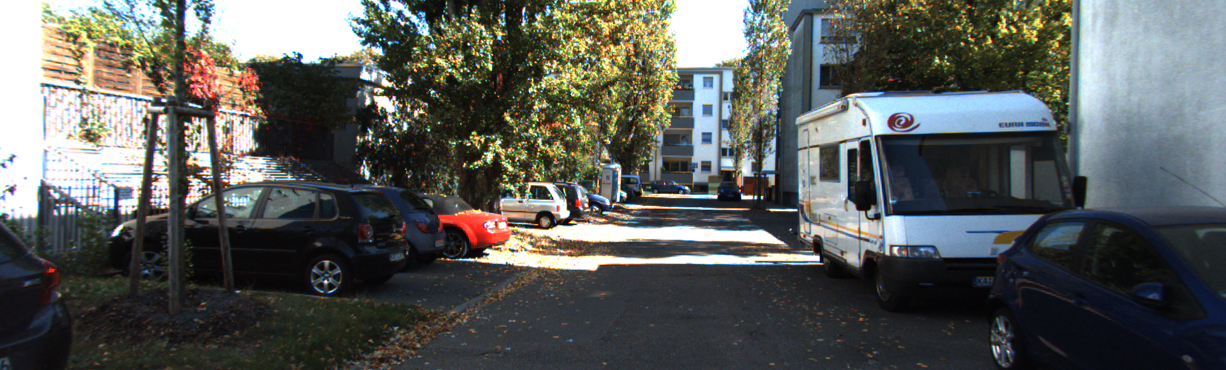} \\
    \rotatebox{90}{\makebox[1.5cm][c]{\scriptsize GT}} &
    \includegraphics[width=\linewidth]{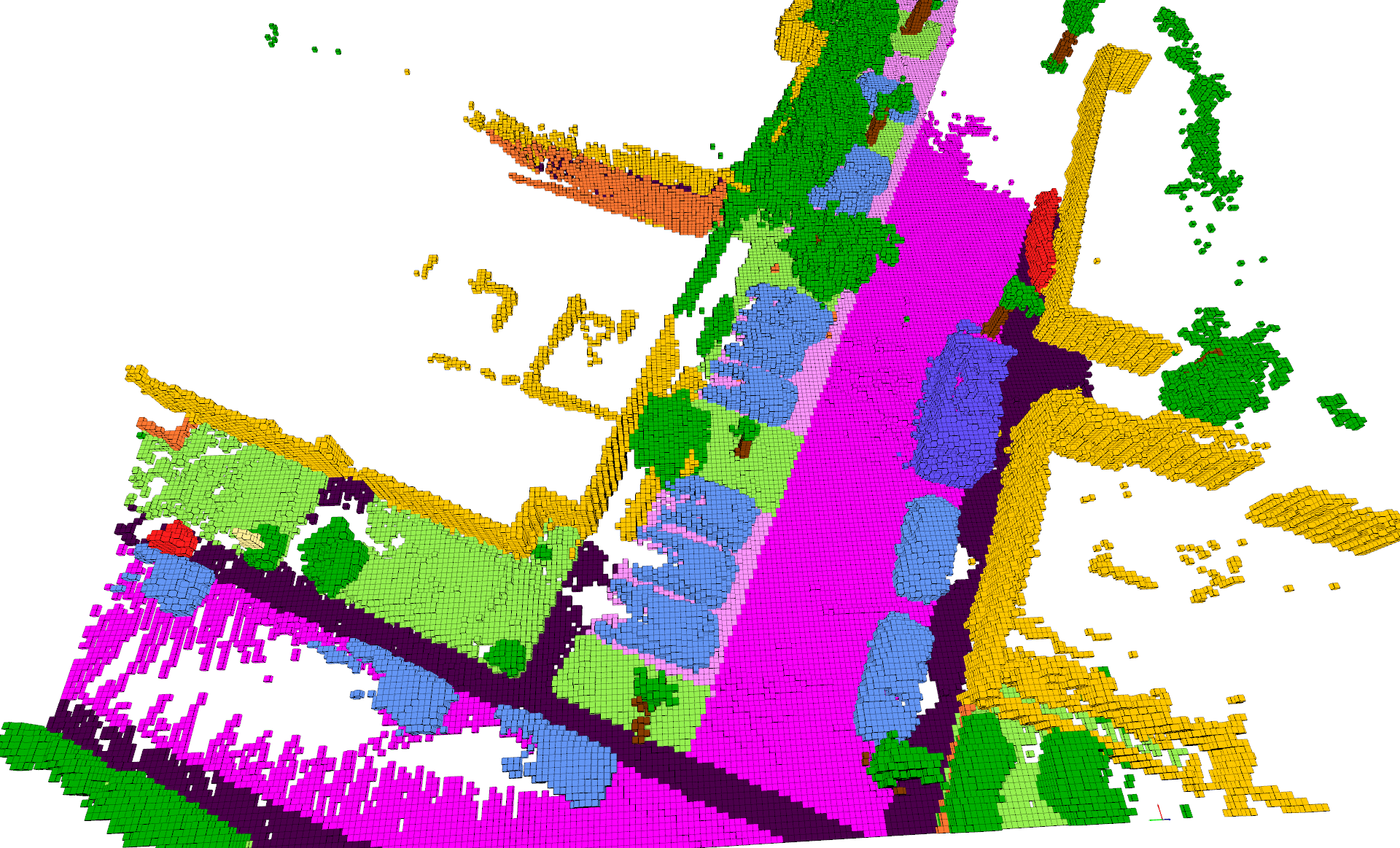}&
    \includegraphics[width=\linewidth]{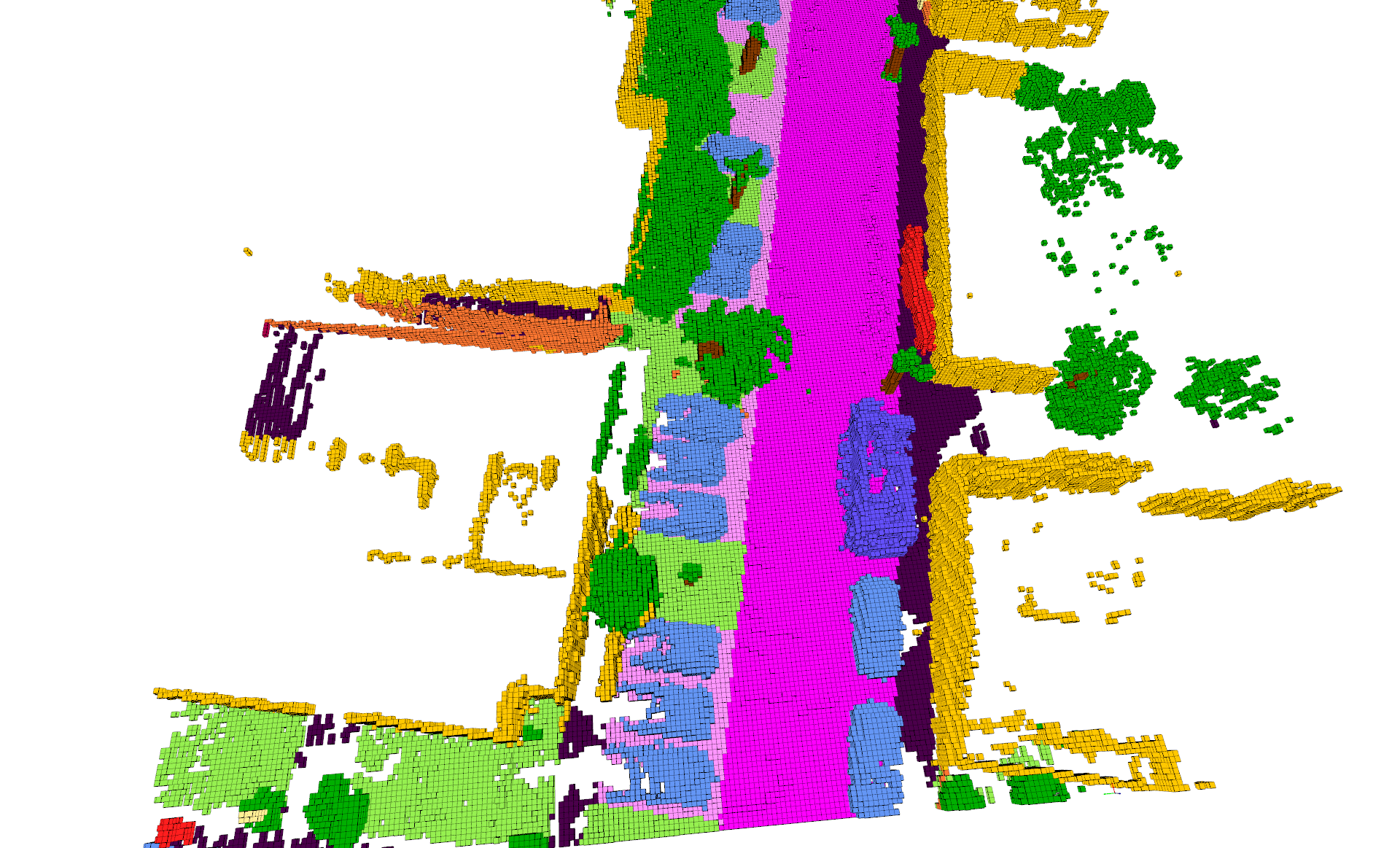} &
    \includegraphics[width=\linewidth]{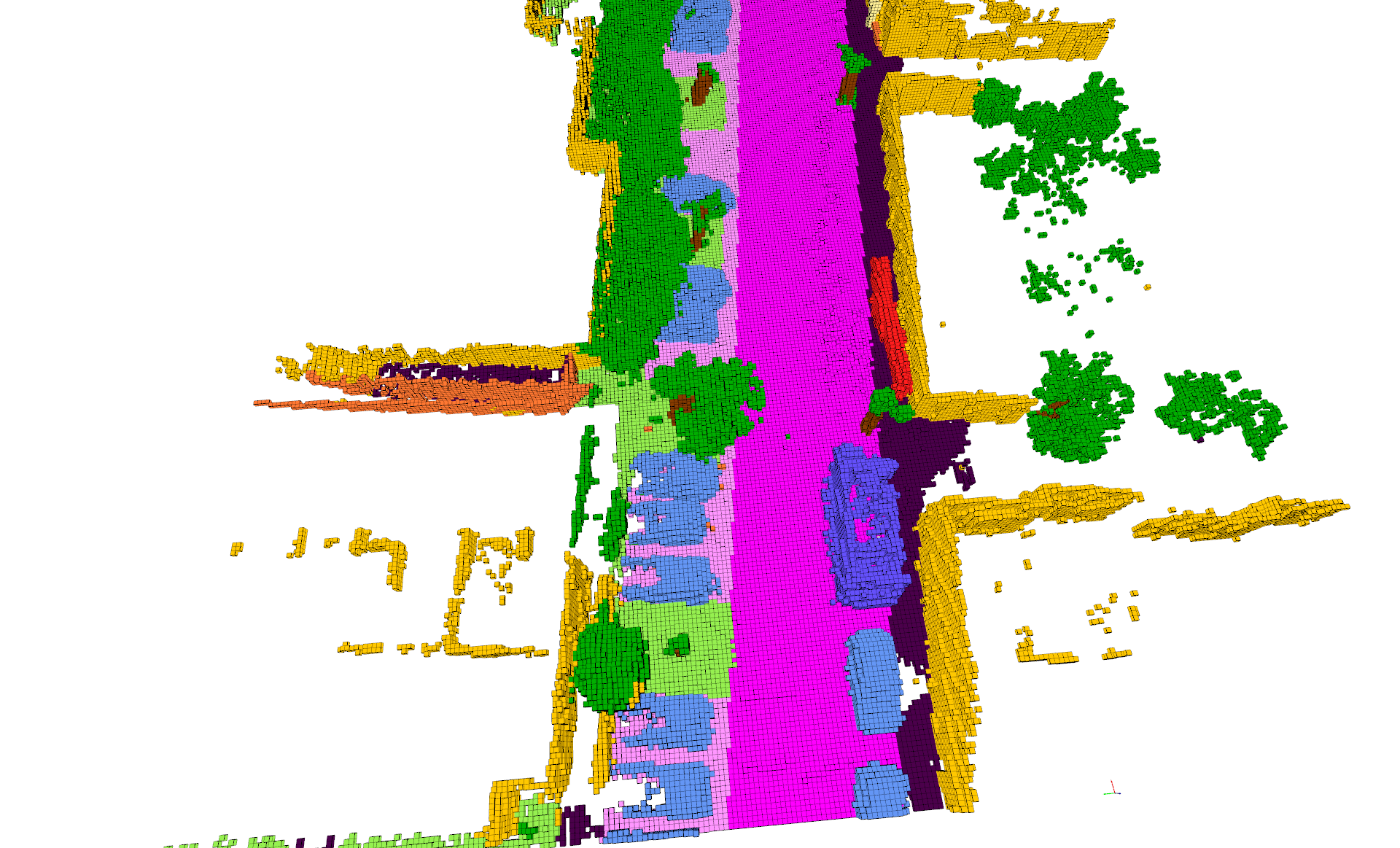}\\

    \rotatebox{90}{\makebox[1.5cm][c]{\scriptsize PDCast}} &
    \includegraphics[width=\linewidth]{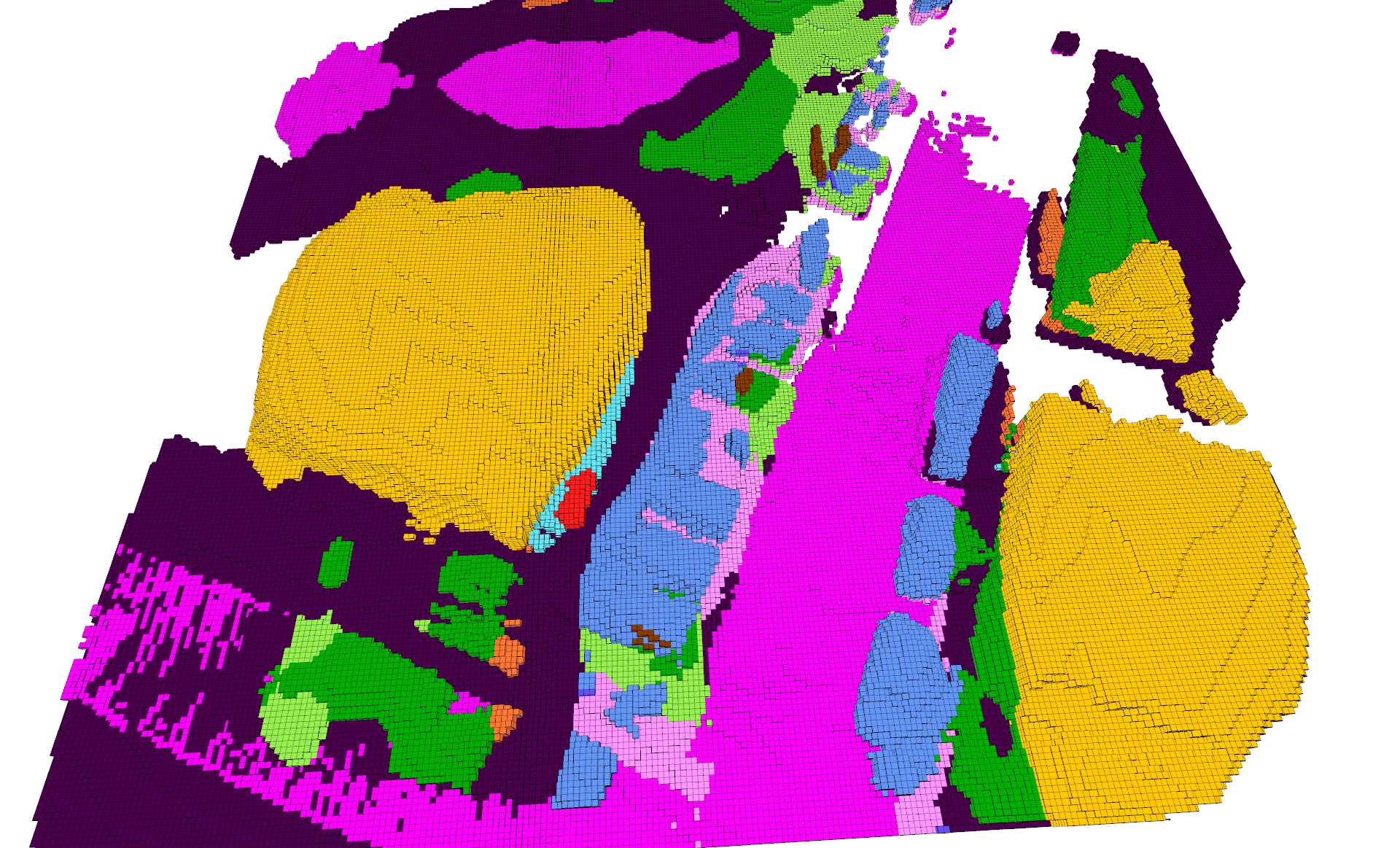}&
    \includegraphics[width=\linewidth]{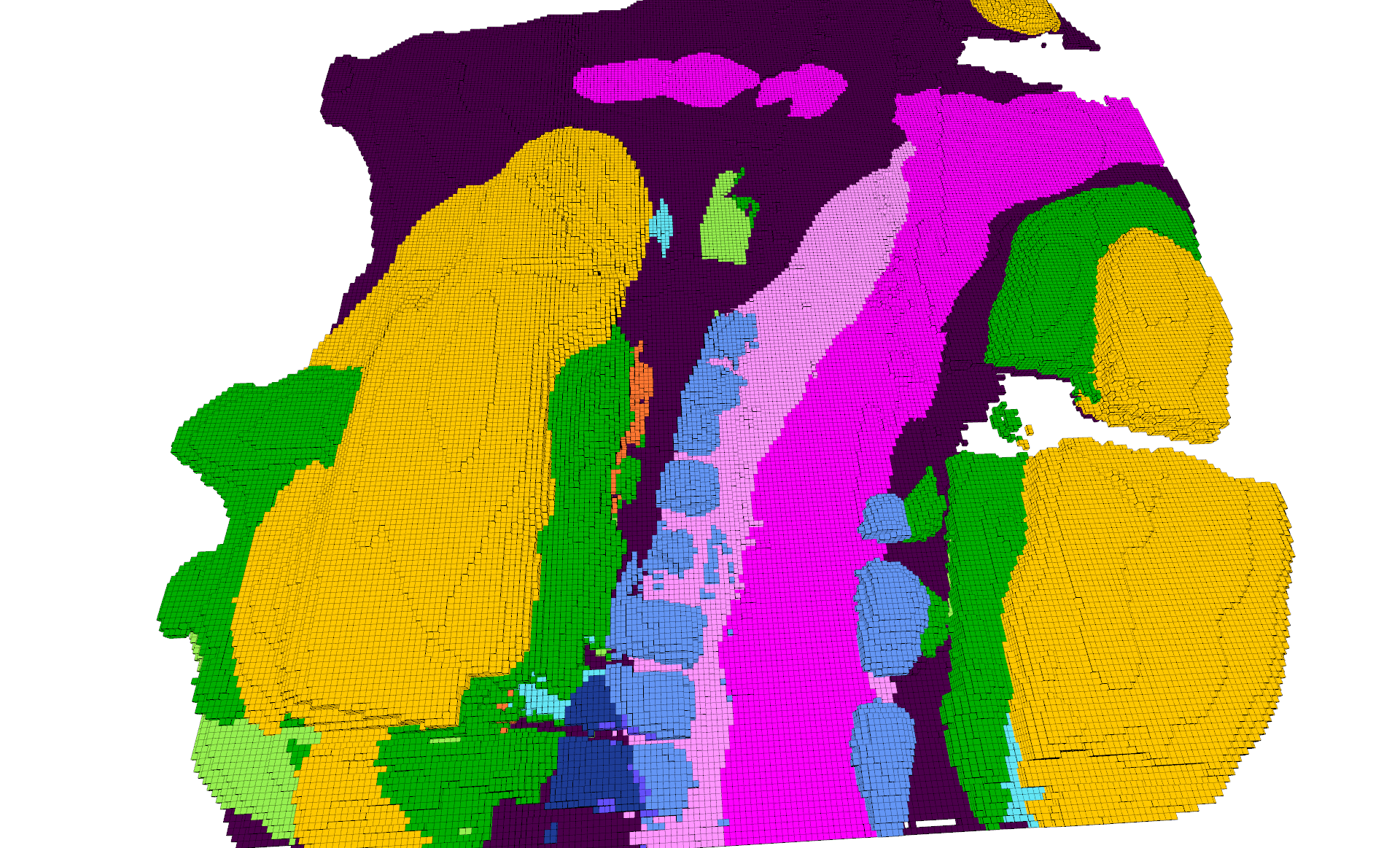} &
    \includegraphics[width=\linewidth]{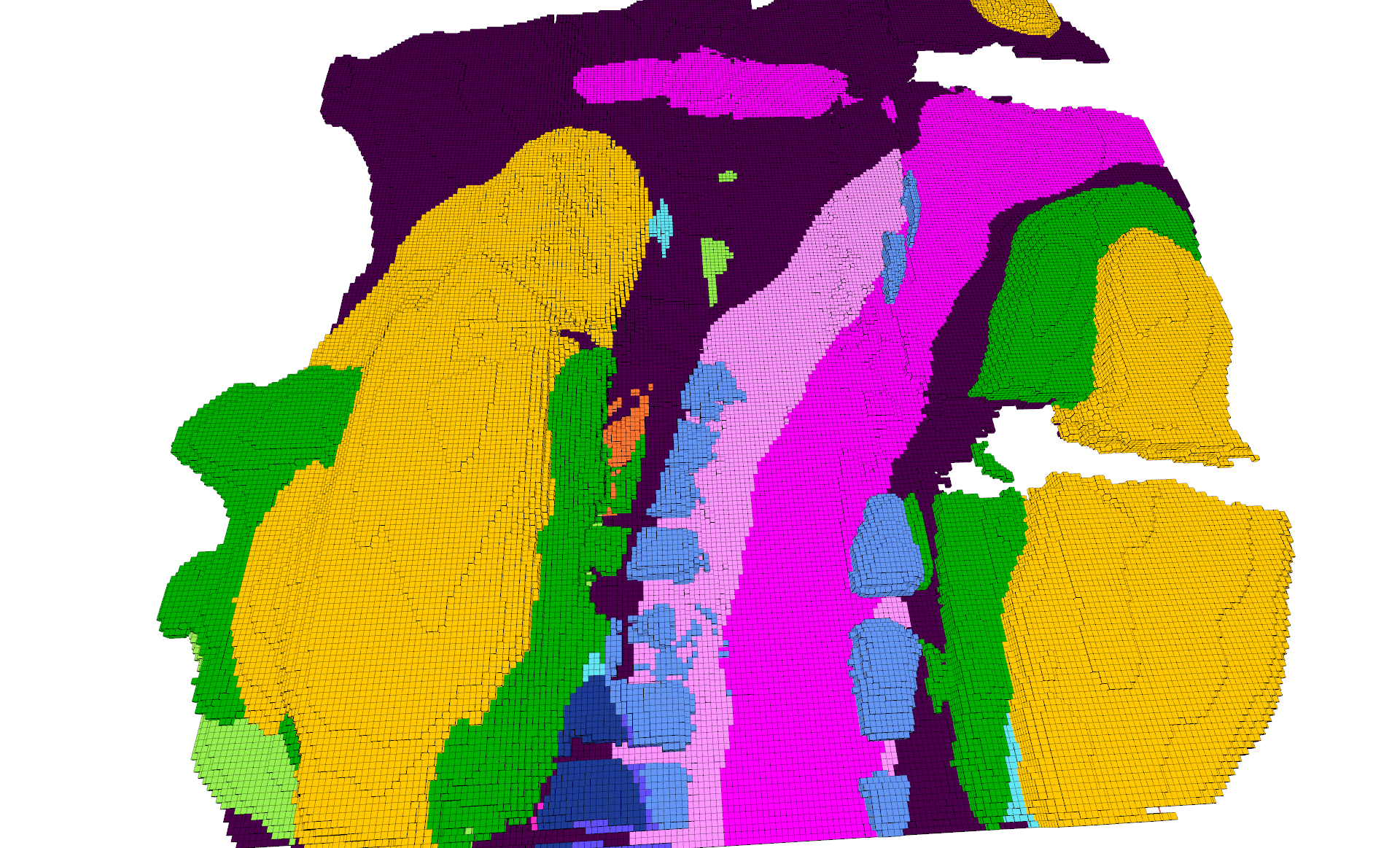}\\
    \rotatebox{90}{\scriptsize \net (ours)} &
    \includegraphics[width=\linewidth]{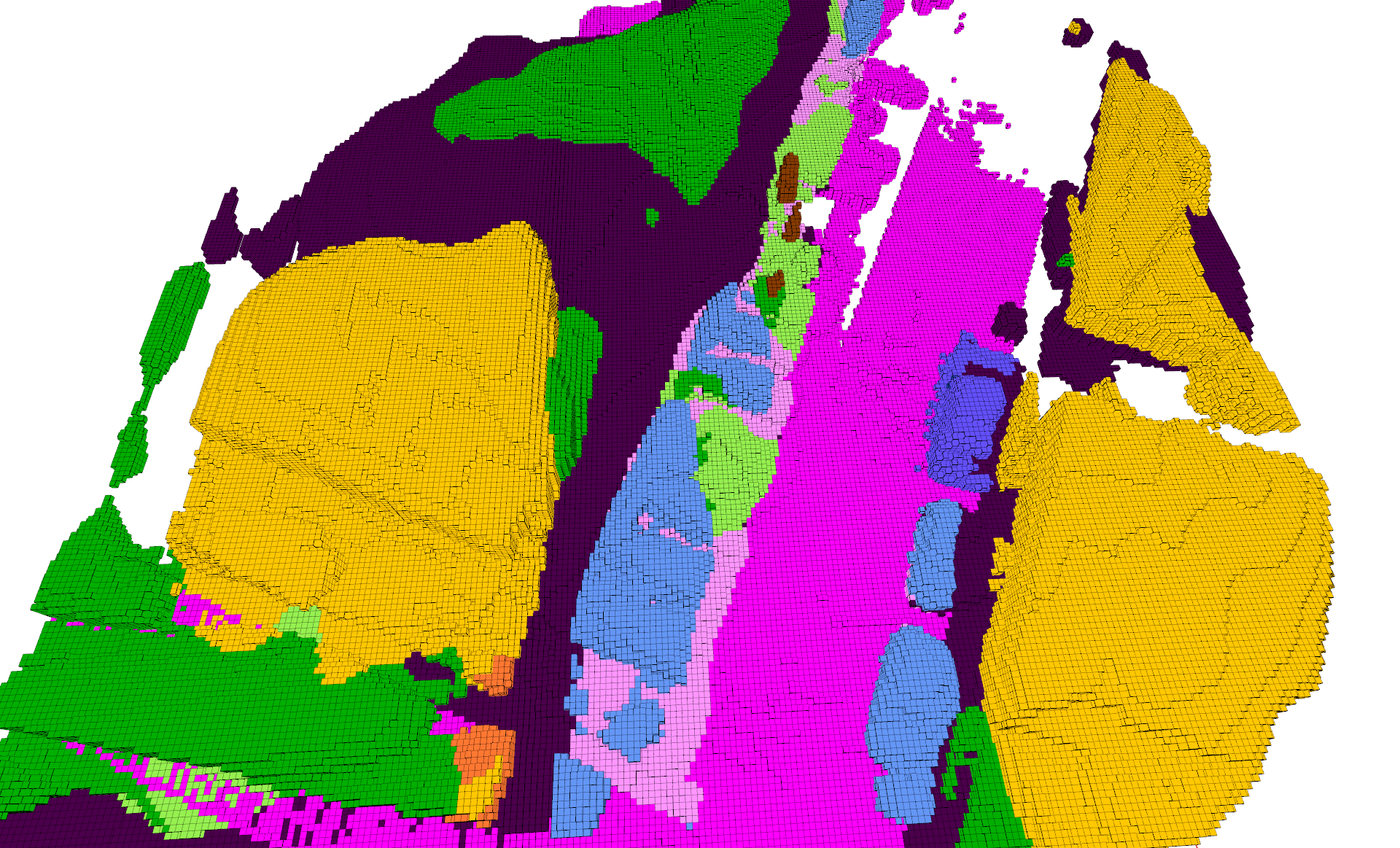}&
    \includegraphics[width=\linewidth]{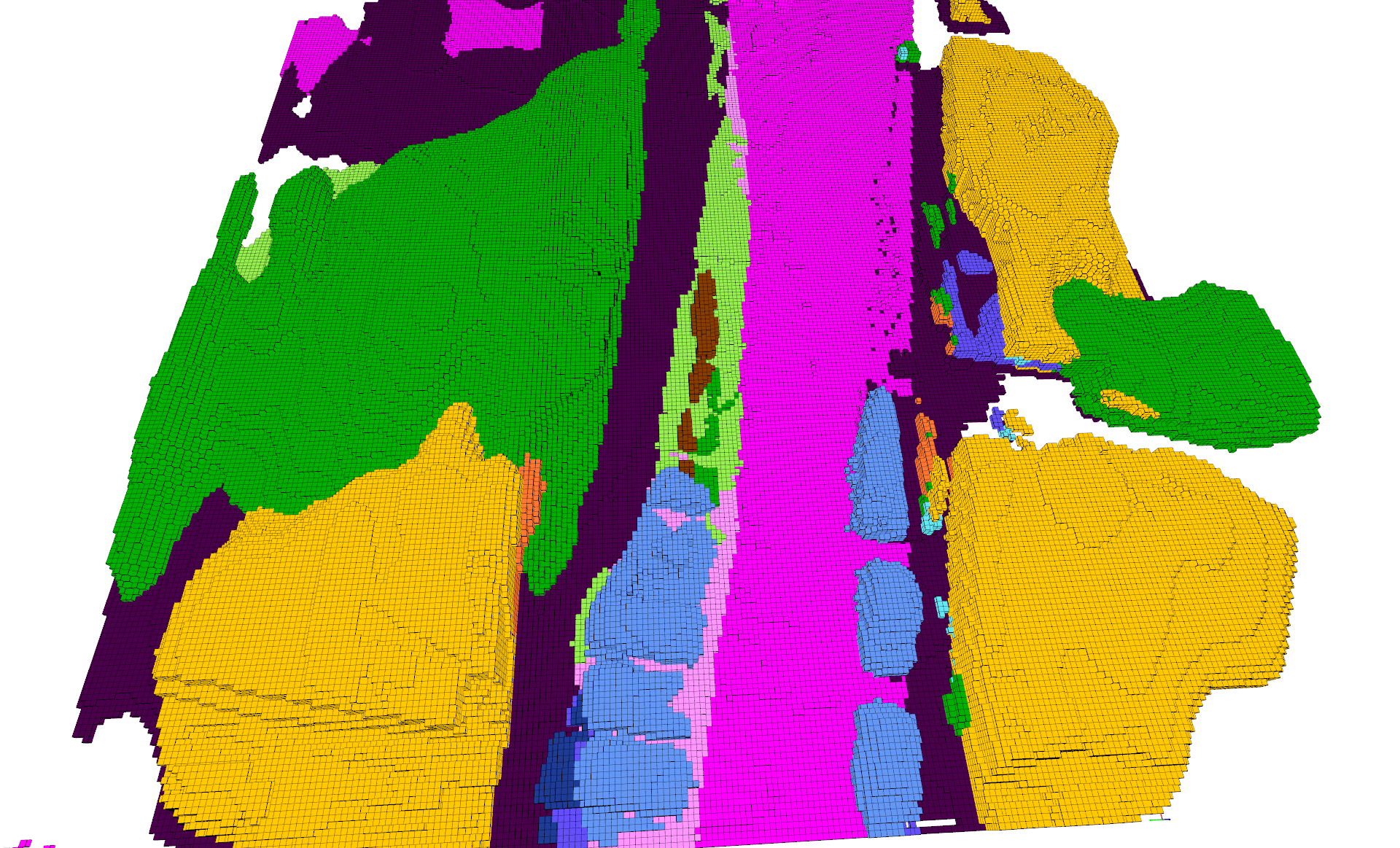} &
    \includegraphics[width=\linewidth]{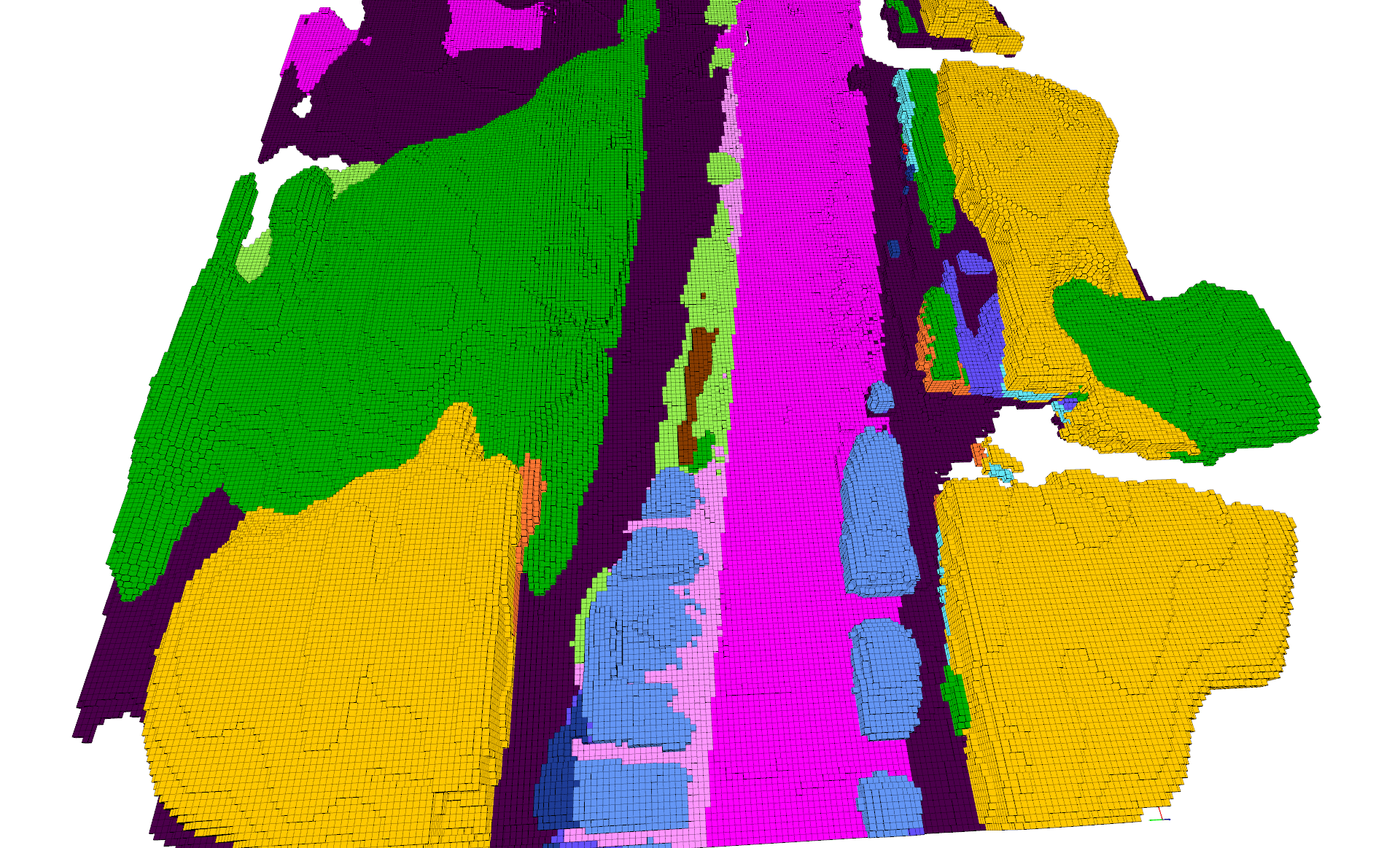}\\
\end{tabular}
\caption{Qualitative comparison of predictions from our proposed \net{} model with the second-best baseline PDCast~\cite{hurtado2024panoptic} on the  SemanticKITTI dataset. We show the image corresponding to the future frame at $T + k$ and the semantic occupancy ground truth (GT).}
\label{fig:qualitative-skitti}
\vspace{-0.5cm}
\end{figure}

\subsubsection{Forecasting module design}
We analyze design choices for the forecasting module and report the results in Tab.~\ref{tab:forecasthead-init} to Tab.~\ref{tab:forecasthead-layers}. Tab.~\ref{tab:forecasthead-init} examines future state query initialization strategies, where deriving the query from current features outperforms the learned baseline (19.68\% vs. 18.36\% mIoU). This suggests that initializing queries with semantically meaningful information from the current frame provides a stronger starting point for forecasting, whereas relying on parameters learned from scratch is less effective. Tab.~\ref{tab:forecasthead-emb} evaluates different configurations of contextual embeddings. Individual temporal, scale, or camera embeddings each improve over the no-embedding case, highlighting the benefit of explicitly tagging the origins of temporal features for the future interaction layer. The combination of all three embeddings yields the highest performance. Finally, Tab.~\ref{tab:forecasthead-layers} studies the number of future interaction layers per temporal interaction. Performance improves as layers increase from one to three (16.87\% to 19.68\% mIoU). Adding a fourth layer offers no further gain, suggesting that three layers achieve the best balance by capturing sufficient temporal dependencies.

\subsection{Qualitative Results}
We qualitatively compare the performance of our model \net{} with $I^{2}$-World and PDCast, the best-performing baselines on Occ3D-nuScenes and SemanticKITTI, respectively. In Fig.~\ref{fig:qualitative-nuscenes}, we observe that both \net{} and $I^{2}$-World effectively forecast large amorphous regions such as roads (pink), vegetation (green), and other flat surfaces, although performance degrades as the prediction horizon increases. At a \SI{1}{\second} horizon, both models successfully forecast the two buses depicted in yellow (one static and one moving), but our approach provides more accurate segmentation. This advantage also extends to a mix of static and dynamic cars (blue) as well as finer man-made structures such as poles and boards (off-white). The trend continues at a \SI{2}{\second} horizon, while at \SI{3}{\second} both models reach their forecasting limits and fail to preserve finer structures. Although they still reliably forecast the static bus and amorphous regions, our model remains slightly more accurate. This \SI{3}{\second} limit highlights a common challenge in forecasting, where accumulated errors and the inherent uncertainty of future states make it increasingly difficult to predict dynamic and fine-grained objects over longer horizons.

In Fig.~\ref{fig:qualitative-skitti}, for semantic occupancy forecasting from a monocular perspective, at the \SI{1}{\second} horizon, both \net{} and PDCast demonstrate encouraging performance. However, our model provides more accurate segmentation for objects closer to the camera's field of view. For example, in the ground truth (GT), the left edge of the road (pink) is lined with buildings, which is accurately captured by our model's forecasting prediction. As we move to \SI{2}{\second} and \SI{3}{\second} horizons, the baseline PDCast incorrectly forecasts a left turn in the road and parking areas (light pink), influenced by the turning motion of the ego car in past and current frames. In contrast, our model maintains a straight-road prediction that is closer to the GT depiction of the scene.  These qualitative comparisons show that \net{} consistently preserves scene geometry and fine structures across both benchmarks. This is enabled by its future state synthesizer that jointly models temporal interactions from past and current states while grounding their evolving dynamics using future state observations during training.\looseness=-1

\section{Conclusion}

In this work, we introduced \net{}, the first vision-based framework for semantic occupancy forecasting. Unlike prior approaches that rely on past occupancy predictions, \net{} directly uses multi-view camera images to jointly forecast future occupancy states and semantic categories. We established a new benchmark on SemanticKITTI and evaluated our method on Occ3D-nuScenes, adapting two 2D forecasting modules as baselines. To tackle this task, we proposed a novel architecture that combines a temporal cross-attention forecasting module with a view transformer, a 3D encoder, and a semantic occupancy head. Extensive experiments and ablations demonstrated that \net{} consistently outperforms both occupancy-input and image-input baselines across multiple horizons and datasets. Our forecasting module proves effective under both multi-view and monocular settings, showing that aligning future-aware image-plane features with observed frames provides strong temporal reasoning and improves semantic occupancy forecasting performance.\looseness=-1

{\footnotesize
\bibliographystyle{IEEEtran}
\bibliography{references}
}

\end{document}